%% file: main_arXiv.tex
\documentclass[a4paper,10pt]{article}
\usepackage[margin=1in]{geometry}
\usepackage[utf8]{inputenc}
\usepackage[T1]{fontenc}
\usepackage{lmodern}
\usepackage{amsmath,amsthm,amssymb,amsfonts}
\usepackage{microtype}
\usepackage{graphicx,wrapfig,booktabs,array}
\usepackage{natbib}
\usepackage{xcolor}
\usepackage{algorithm}
\usepackage[noend]{algpseudocode}
\usepackage{tcolorbox}
\tcbuselibrary{breakable,skins}
\usepackage{pgfplots}
\pgfplotsset{compat=1.18}
\usepackage{titletoc}
\usepackage{hyperref}
\hypersetup{colorlinks=true,linkcolor=blue,citecolor=blue,urlcolor=blue}
\IfFileExists{command.tex}{\input{command}}{}
\providecommand{\E}{\mathbb{E}}
\providecommand{\ind}{\mathbf{1}}
\providecommand{\KL}{\operatorname{KL}}
\providecommand{\pos}[1]{\left[#1\right]_{+}}
\makeatletter
\@ifundefined{theorem}{\newtheorem{theorem}{Theorem}}{}
\@ifundefined{corollary}{\newtheorem{corollary}{Corollary}}{}
\makeatother
\definecolor{DEOPurple}{HTML}{75549A}
\definecolor{DEOTeal}{HTML}{428B94}
\definecolor{DEORuntimeRed}{HTML}{C55D68}
\definecolor{DEORuntimeGreen}{HTML}{278A78}
\definecolor{DEOPromptBlue}{HTML}{244F9A}
\definecolor{DEOPromptBack}{HTML}{F3F5FB}
\floatstyle{plain}
\restylefloat{algorithm}
\algrenewcommand{\algorithmicindent}{1.2em}
\algrenewcommand{\alglinenumber}[1]{\normalfont\scriptsize #1:}
\newcommand{\DEOnote}[1]{%
  \Statex\hspace{\algorithmicindent}%
  {\footnotesize\color{black!55}$\triangleright$~#1}%
}
\newenvironment{DEOalgorithm}[2]{%
  \begin{tcolorbox}[colback=black!1,colframe=black!25,
    boxrule=0.5pt,arc=1.5pt,left=6pt,right=6pt,top=6pt,bottom=6pt,
    before skip=0pt,after skip=0pt]
  \footnotesize\refstepcounter{algorithm}\label{#2}
  \noindent\textbf{Algorithm \thealgorithm:
    \textcolor{DEOPurple}{#1}}\par\smallskip
}{\end{tcolorbox}}
\newtcolorbox{DEOprompt}[1]{
  enhanced,breakable,sharp corners,
  colback=DEOPromptBack,colframe=DEOPromptBlue,
  colbacktitle=DEOPromptBlue,coltitle=white,
  fonttitle=\small\bfseries,fontupper=\small,
  title={#1},title after break={#1 (continued)},
  boxrule=0.6pt,left=7pt,right=7pt,top=6pt,bottom=6pt,
  before skip=6pt,after skip=7pt,
  before upper={\setlength{\parindent}{0pt}\setlength{\parskip}{0pt}\raggedright}
}
\title{\bfseries Direct Self-Evolving Optimization: Evolving LLMs without Challenger Training}
\author{%
  {\normalsize Yuyang Deng\textsuperscript{1}\qquad
  Yu Wang\textsuperscript{1}\qquad
  Jiayun Wang\textsuperscript{2}}\\[0.5em]
  {\small\textsuperscript{1}Accenture, Center for Advanced AI}\\
  {\small\textsuperscript{2}Georgia Institute of Technology}\\[0.4em]
  {\footnotesize\texttt{matthewgo2009@gmail.com}\quad
  \texttt{feather1014@gmail.com}\quad
  \texttt{pjwang@gatech.edu}}%
}
\date{}

\begin{document}
\maketitle

\begin{abstract}
Self-evolving language models improve by generating tasks and learning from
their own feedback, but adapting the task generator often requires a
separate challenger-training loop. Can we generate tasks adapted to the
current solver without explicitly training a challenger? We introduce
\textbf{D}irect Self-\textbf{E}volving \textbf{O}ptimization (DEO), which
replaces challenger parameter updates with solver-guided task sampling.
The KL-regularized challenger objective defines an exponential tilt of a
fixed base task distribution. DEO uses this distribution as a sampling
target: a frozen LLM generates and mutates tasks, the solver scores them,
and an approximate Metropolis selection rule refines the training pool.
Only the solver is trained. Theoretically, for an idealized variant that
samples exactly from the tilted distribution, and under regularity, local
gradient-dominance, and initialization conditions, we show that
DEO learns distributionally robust reasoning ability. In experiments, DEO achieves reasoning
performance competitive with R-Zero while using over $50\%$ less wall-clock training time, and improves reasoning accuracy over
a no-walk ablation. Replacing the task generator with a frozen API-only
LLM further improves the local solver, illustrating a capability enabled
by removing challenger training.
\end{abstract}

\section{Introduction}

Self-evolving training offers a way to improve large language models (LLMs)
by generating their own tasks and learning from the resulting feedback.
Recent methods, including R-Zero~\citep{huang2025r}, Absolute Zero Reasoner
(AZR)~\citep{zhao2025absolute}, and Agent0~\citep{xia2025agent0}, couple
task proposal with task solving to construct an adaptive curriculum.
A challenger seeks tasks that challenge the current solver, and the solver
learns from the generated tasks during this interaction. R-Zero and
Agent0 train separate models for these roles; AZR trains one model to
perform both (but still two-stage training). These approaches demonstrate that pretrained models can
improve without an externally supplied task--answer dataset for this
stage of training.

Training the challenger, however, adds another optimization process to every
self-evolving round. A separately trained challenger also requires its
own training state and access to trainable model parameters. Yet the
solver ultimately consumes the tasks produced by the challenger, rather
than the challenger's parameters. This motivates our central question:
\begin{quote}
\emph{Can we directly generate tasks lying in the current solver's capability frontier without
explicitly training a challenger?}
\end{quote}

Our starting point is to view the challenger as a distribution over tasks.
For a fixed solver, maximizing challenger reward with a KL penalty to a
base task distribution has an exponential-tilt solution. This is a
standard property of KL-regularized optimization~\citep{rafailov2023direct}:
tasks with higher challenger reward receive greater relative probability.
The closed form specifies a target distribution; it does not by itself
make sampling tractable. It nevertheless suggests an alternative to
learning a parameterized challenger: adapt the task samples directly.
\emph{An adaptive task distribution need not require updates to the
generator's parameters.}
\begin{figure}[t]
    \centering
    \includegraphics[width=\linewidth]{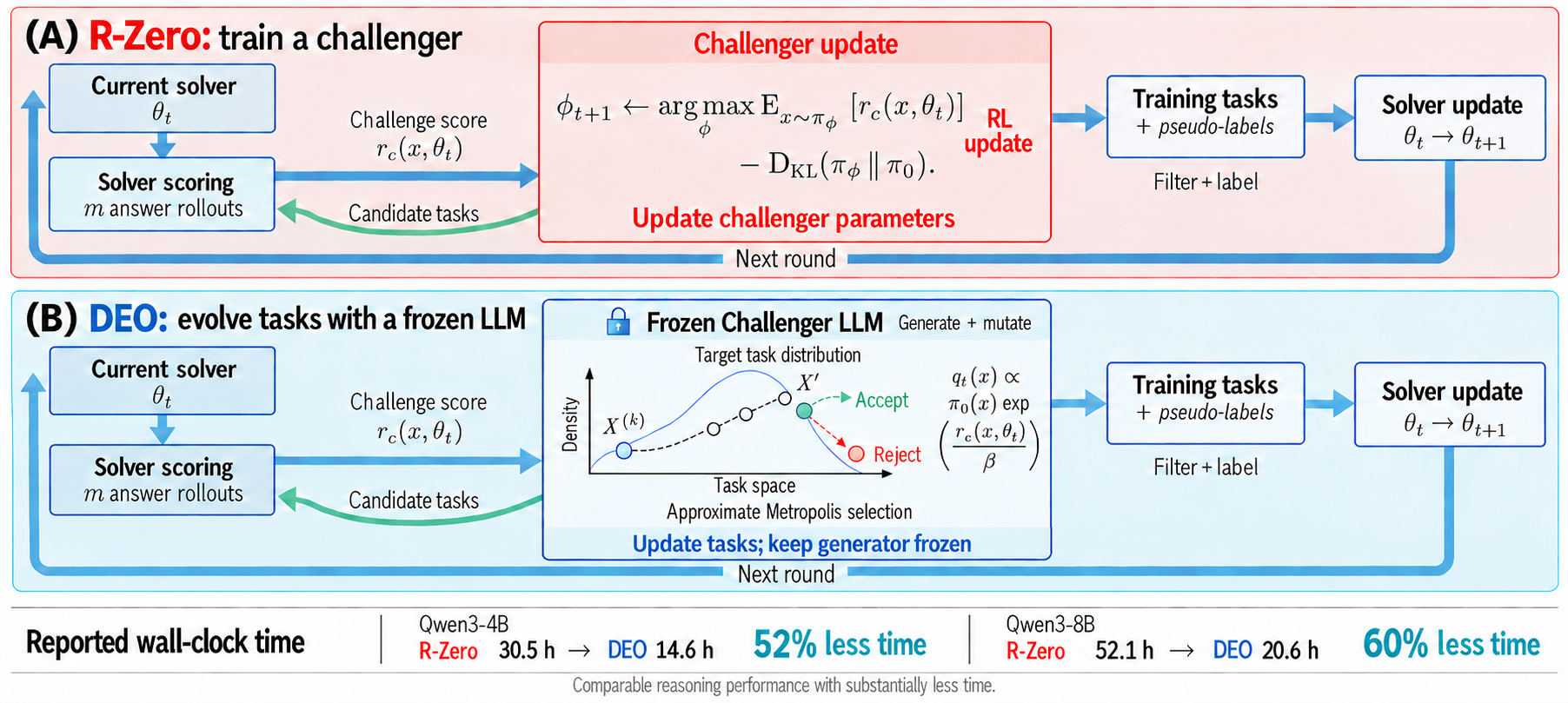}
    \caption{Comparison between traditional challenger training self-evolving method and our DEO.  (A) Conventional
    self-evolving method updates both a task proposer and a solver. (B) DEO keeps the
    generator frozen and uses solver feedback to select task mutations;
    only the solver receives parameter updates. DEO uses more than half less wall-clock time than R-Zero while yielding competitive performance. }
    \label{fig:algorithm_flow}
\end{figure}


We develop this idea into \textbf{D}irect Self-\textbf{E}volving
\textbf{O}ptimization (DEO). A frozen LLM generates an initial task pool
and proposes mutations conditioned on the current tasks. The solver
evaluates these proposals, and a Metropolis-style selection rule guides
the pool toward the current challenger target: if the mutated task looks more challenging to the current solver, we tend to accept it. Through this LLM-based mutation, we produce tasks that lie near the frontier of the current solver's capability. We then use the mutated tasks and
solver-produced pseudo-labels to train the solver. Drawing on Metropolis--Hastings
sampling~\citep{metropolis1953equation,hastings1970monte}, this procedure
uses LLM inference to adapt the curriculum while eliminating challenger
parameter updates. Because the generator only needs inference access, DEO can
also use a closed-source LLM as the task proposer and mutator.

We then ask, {\em what does DEO actually learn?} We analyze an idealized
variant of DEO that samples exactly from the optimal challenger distribution.
The analysis shows that DEO learns a model whose
worst-case answer uncertainty over a KL neighborhood of the base task
distribution is small (Theorem~\ref{thm:optimal-challenger}). If the initial model's
modal answers are mostly correct, this worst-case answer concentration translates into a distributionally robust answer-accuracy
bound (Corollary~\ref{cor:accuracy}). These results explain the role of adversarial task
sampling: it learns {\em distributionally robust reasoning ability}. They do not, however, imply uniform superiority on every evaluation domain.

We conduct experiments on mathematical reasoning. DEO
remains competitive with R-Zero, which trains a challenger, while using $52\%$--$61\%$ less wall-clock time. Because DEO replaces challenger parameter updates with LLM inference and selection, it can also use a closed-source model as generator and mutator: replacing the open-source generator with Claude Haiku 4.5 further improves the
solver's reasoning ability. Together, these
results support solver-guided task sampling as a practical alternative
to explicit challenger training.

\section{Related Work}
\label{sec:related-work}
\paragraph{Self-evolving training.}
Self-generated data and feedback provide several routes to improving LLM
parameters. Self-Instruct~\citep{wang2022selfinstruct} generates and filters
instruction data for fine-tuning, while STaR~\citep{zelikman2022star}
iteratively trains on generated rationales that produce known correct
answers. Evol-Instruct~\citep{xu2023wizardlm} rewrites seed instructions
into more complex training examples, and Self-Rewarding Language
Models~\citep{yuan2024selfrewarding} use model-generated judgments for
iterative preference optimization. These methods reduce the need for
manual supervision, with different requirements for seed tasks, labels,
and feedback.

An interacting-agent perspective on LLM training is discussed
by \citet{liu2024large}. Closest to our setting, R-Zero~\citep{huang2025r},
AZR~\citep{zhao2025absolute}, and Agent0~\citep{xia2025agent0} adapt task
generation to the evolving solver. R-Zero trains separate challenger and
solver models; AZR trains a single model to propose and solve
executor-verifiable tasks; and Agent0 co-evolves curriculum and executor
agents with tool use. DEO shares their goal of constructing an adaptive
curriculum, but replaces proposer parameter updates with solver-guided
task sampling. LLM-based task rewriting itself has a precedent in
Evol-Instruct; DEO instead grounds task selection in a regularized
challenger objective and uses the current solver's evaluations to guide
Metropolis-style refinement toward its induced task distribution.

\paragraph{Inference-time refinement and evolution.}
A complementary line of work improves outputs through repeated LLM
inference. Self-Refine~\citep{madaan2023selfrefine} alternates answer
generation, self-feedback, and revision, while
Reflexion~\citep{shinn2023reflexion} stores verbal reflections to guide
subsequent attempts without updating model weights. Tree of
Thoughts~\citep{yao2023tree} searches over intermediate reasoning steps.
Evolutionary approaches include Mind
Evolution~\citep{lee2025evolving}, which generates, recombines, and refines
candidate responses, and FunSearch~\citep{romeraparedes2024mathematical},
which pairs LLM-generated program variants with an automated evaluator.
AlphaEvolve~\citep{novikov2025alphaevolve} extends this approach to
larger programs and algorithmic discovery, using LLM-proposed code
edits, automated evaluators, and a population of candidate solutions.
Its search optimizes programs against user-specified evaluation metrics.
Related prompt optimization methods such as
GEPA~\citep{agrawal2025gepa} use reflections on execution traces to evolve
system prompts. DEO likewise uses a frozen LLM to propose revisions and
feedback to select them, but the evolving objects are training tasks.
Its walk constructs a curriculum for subsequent solver parameter
updates, connecting inference-based search to self-evolving training.

\section{Direct Self-Evolving Optimization}
\label{sec:framework}
\label{sec:method} 
Existing self-evolving methods typically train two roles: a challenger (proposer) that generates a pool of training tasks, and a solver that tries to solve them. The challenger is trained to generate tasks that are more challenging for the solver, while the solver is trained to solve these tasks as correctly as possible.

\subsection{From challenger training to a task-sampling target}

Let $\pi_{\theta}$ be the solver policy parameterized by $\theta\in\R^d$, and let $\pi_0(x)$
be a fixed base task distribution induced by a frozen generator and its prompting
protocol. Let $r_c(x,\theta )$ score the
challenging-ness of  task $x$ to the current solver $\pi_{\theta}$. In the existing framework such as~\cite{huang2025r}, the challenger training stage's objective is 
\begin{align}
  \max_{\phi} \left\{\E_{x\sim \pi_{\phi}}r_c(x,\theta_t)
                    -\beta\KL(\pi_{\phi}\|\pi_{0})\right\},\label{eq:raw-challenger-obj}
\end{align}
where $\phi$ denotes the parameters of the challenger model (an LLM), $\theta_t$ is the solver at round $t$, and the KL term with coefficient
$\beta>0$ controls departure from the base task law.
In words, the challenger is trained to place more probability on tasks with higher challenge scores. However, training this model can be costly. Inspired by DPO~\citep{rafailov2023direct}, we consider the following distributional formulation of~\eqref{eq:raw-challenger-obj}:
\begin{equation}
    q_t\in\arg\max_{q\ll\pi_0}
       \left\{\E_{x\sim q}r_c(x,\theta_t)
                    -\beta\KL(q\|\pi_0)\right\},
    \label{eq:challenger-obj}
\end{equation}

The key observation is that $q_t$ admits the following closed form:
\begin{equation}
    q_t(x)=\frac{\pi_0(x)\exp(r_c(x,\theta_t)/\beta)}{Z_{\theta_t}},
    \qquad Z_{\theta_t}=\E_{x\sim\pi_0}\exp(r_c(x,\theta_t)/\beta).
    \label{eq:closed-form-q}
\end{equation}
This follows from the Gibbs variational identity~\citep{donsker1975asymptotic}. Thus $q_t$ reweights the base distribution, giving exponentially more mass to tasks that are more challenging for the current solver $\theta_t$. As the solver changes, the
reward changes and hence $q_t$ changes. If we can directly draw tasks from $q_t$, then we are able to obtain useful training samples 
without fitting a new generator at each round.

\subsection{LLM-based Metropolis--Hastings sampling}
However, directly sampling from $q_t$ is intractable. We thus propose a sampling step based on \emph{LLM-based mutation and Metropolis--Hastings-style
selection}, LLM-MH for short. A frozen LLM (the generator inducing $\pi_0$, or even a closed-source model) first generates a seed pool. It then proposes
revisions conditioned on each current task, while the frozen current
solver evaluates the proposals. A Metropolis--Hastings-style selection rule uses
these evaluations to accept or reject each revision.

In detail, each round starts with $n_t$ seed tasks from the frozen LLM.
For each task $x$, we prompt the same LLM to produce a structurally
different task $x'$ by some invertible operation:
\begin{align*}
    x' \sim \pi_{0} (\cdot |x, \text{mutation prompt})
\end{align*}
Denote this prompted proposal kernel by
$K_{\mathrm{LLM}}(x'\mid x)$. Reversible task edits can make reverse
moves possible, but do not guarantee symmetric proposal probabilities
or reversibility with respect to $\pi_0$.

The current solver then draws $m$ fresh responses to $x'$. We accept $x \leftarrow x'$ with probability
\begin{equation}
  \hat \alpha(x,x')=\min\left\{1,
      \exp\left(\frac{ \hat r_c(x',\theta_t)
                          -  \hat r_c(x,\theta_t)}{\beta}\right)\right\}.
    \label{eq:task-metropolis-acceptance}
\end{equation}
where $\hat r_c$ estimates $r_c$ from $m$ solver rollouts. This is an
approximate selection rule. Exact Metropolis--Hastings targeting $q_t$
would use exact rewards and also include the factor
\[\frac{\pi_0(x')K_{\mathrm{LLM}}(x\mid x')}{
\pi_0(x)K_{\mathrm{LLM}}(x'\mid x)}.\] This factor cancels only for a
$\pi_0$-reversible proposal. The practical walk omits this intractable
factor and uses noisy scores and finitely many steps. The exact-sampling
assumption in Section~\ref{sec:theory} is therefore an idealization,
not a claimed consequence of running this walk longer. In our experiments,
$\widehat r_c$ is a triangular agreement score with a batch-level BLEU
repetition penalty, so acceptance is evaluated on the whole batch
(Appendix~\ref{app:llm-walk}).

\subsection{Solver learning and the self-evolving loop}

For each generated task, obtain $m$ responses from the current solver and
form a pseudo-label $\widehat z_{t,i}=\mathcal A_m(y_{t,i,1:m})$, where
$\mathcal A_m$ is an aggregation rule. The solver remains frozen during
task generation and labeling. Supervision is derived from these responses,
not from answers supplied by the task generator.  

Let $r_s(x,\theta;z)$ denote the solver loss against
fixed supervision $z$. Throughout, $r_c$ is maximized and $r_s$ is minimized.
Given $\mathcal D_t=\{(x_{t,i},\widehat z_{t,i})\}_{i=1}^{n_t}$, define
\begin{align}
    \widehat L_t(\theta)
       &:=\frac1{n_t}\sum_{i=1}^{n_t}r_s(x_{t,i},\theta;\widehat z_{t,i}),\quad 
    \widehat J_t(\theta) :=\widehat L_t(\theta)
            +\tau_t\mathcal R_s(\pi_\theta,\pi_{\mathrm{ref},t}),
       \label{eq:general-solver-obj}
\end{align}
where the solver regularizer, reference, and coefficient schedule are
independent choices from the challenger's regularization. The goal of solver training is to optimize~\eqref{eq:general-solver-obj} and find $\theta_{t+1}$ that satisfies
\begin{equation}
    \widehat J_t(\theta_{t+1})
       \leq\inf_\theta\widehat J_t(\theta)+\varepsilon_{\mathrm{opt},t}.
    \label{eq:empirical-solver-update}
\end{equation}
where $\varepsilon_{\mathrm{opt},t}$ is the optimization error.

Algorithm~\ref{algorithm:mcmc-deo} gives the DEO loop.
After each update, $\theta_{t+1}$
defines the target and supervision for the next round. The generator
requires only inference access throughout.

\begin{figure}[t]
\centering

\begin{DEOalgorithm}
  {Direct Self-Evolving Optimization (DEO)} 
 {algorithm:mcmc-deo}

\noindent
\textbf{Input:} initial solver $\theta_0$,
frozen generator $\pi_0$, mutation kernel $K_{\mathrm{LLM}}$,
rounds $T$, pool sizes $\{n_t\}$, sweeps $L_{\mathrm w}$, temperature $\beta$,
rollouts $m$, aggregation $\mathcal A_m$.
\par
\noindent
\textbf{Output:} evolved solver $\theta_T$.

\newcommand{\DEOstep}[1]{%
  \State\parbox[t]{\dimexpr\linewidth-\csname ALG@thistlm\endcsname-1pt\relax}{%
    \raggedright #1\strut}%
}
\renewcommand{\DEOnote}[1]{%
  \Statex\hspace*{\csname ALG@thistlm\endcsname}%
  \parbox[t]{\dimexpr\linewidth-\csname ALG@thistlm\endcsname-1pt\relax}{%
    \raggedright\footnotesize\color{black!55}$\triangleright$~#1\strut}%
}
\par\medskip
\noindent
\begin{minipage}[t]{0.48\linewidth}
\vspace{0pt}
\raggedright
\textbf{Main routine}\par\smallskip
\begin{algorithmic}[1]
  \For{$t=0,\ldots,T-1$}
    \DEOstep{Draw seed task pool with $n_t$ tasks from frozen base model: { $S_t \sim \pi_0^{n_t}$}.}
\DEOstep{{ 
      $S_t\leftarrow$ {\color{DEOTeal} LLM-MH}
    ($S_t,\pi_{\theta_t},K_{\mathrm{LLM}},\beta,L_{\mathrm w},m$)}.}
\DEOnote{Approximate the challenger target $q_t$;}

    \For{$i=1,\ldots,n_t$}
      \DEOstep{Draw $m$ responses
        $y_{t,i,1:m}$ from
        $\pi_{\theta_t}(\cdot\mid x_{t,i})$.}
        \DEOstep{$\widehat z_{t,i}
        \leftarrow\mathcal A_m(y_{t,i,1:m})$.}
\EndFor

    \DEOstep{$\mathcal D_t\leftarrow
      \{(x_{t,i},\widehat z_{t,i})\}_{i=1}^{n_t}$.}
\DEOstep{{ 
      Train $\theta_{t+1}$ by minimizing
      $\widehat J_t$ in \eqref{eq:general-solver-obj}.}}
\EndFor

  \DEOstep{\Return $\theta_T$.}
\end{algorithmic}

\end{minipage}\hfill
\begin{minipage}[t]{0.48\linewidth}
\vspace{0pt}
\raggedright
\textbf{Sampling subroutine}\par\smallskip
\begin{algorithmic}[1]
  \Function{\color{DEOTeal} LLM-MH}
    {$S,\pi_\theta,K_{\mathrm{LLM}},\beta,L_{\mathrm w},m$}
    \DEOstep{Estimate $\widehat r_c(x_i,\theta)$ for each $x_i\in S$
      using $m$ solver responses.}
\For{$\ell=1,\ldots,L_{\mathrm w}$}
      \For{$i=1,\ldots,|S|$}
        \DEOstep{{ 
          $x'\sim K_{\mathrm{LLM}}(\cdot\mid x_i)$}.}\vspace{-2mm}
\DEOnote{Prompt the frozen LLM
          to mutate the current task.}

          \DEOstep{Draw $m$ responses from
            $\pi_\theta(\cdot\mid x')$.}
\DEOstep{Compute $\Delta\leftarrow
            \widehat r_c(x',\theta)
            -\widehat r_c(x_i,\theta)$ }\vspace{-3mm}
\DEOnote{Finite roll-out estimation of the  challenge score.}

          \DEOstep{$u\sim\operatorname{Uniform}(0,1)$.}
\If{$u\leq\min\{1,\exp(\Delta/\beta)\}$} 
          \DEOstep{Replace $x_i$ in $S$ by $x'$ and cache its score.}
\EndIf

      \EndFor
    \EndFor

    \DEOstep{\Return $S$.}
\EndFunction
\end{algorithmic}

\end{minipage}

\end{DEOalgorithm}
\end{figure}

\section{Theoretical Analysis: Idealized DEO Learns Distributionally Robust Reasoning}
\label{sec:theory}
\label{sec:concentration-instance}
What does the model learn from DEO? We answer this question by deriving a generalization bound for DEO. For ease of analysis, we assume that the tasks are sampled independently from
the exact optimal challenger $q_t$ instead of approximately sampling from it using LLM mutation and Metropolis selection. That is, at iteration $t$, $S_t \sim q_t^{n_t}$.
The pseudo-labels are still generated by the current solver $\theta_t$ via finite rollout and majority voting. Let $P_\theta(a\mid x)$ be the canonical-answer
probability, $p_\theta(x)=\max_aP_\theta(a\mid x)$, and $a_t(x)$ the
population modal answer of solver $\theta_t$. Instantiate
\begin{equation}
    r_c(x,\theta)=1-p_\theta(x),\qquad
    r_s(x,\theta;z)=1-P_\theta(z\mid x).
    \label{eq:concentration-rewards}
\end{equation}
The empirical solver uses
\begin{equation}
    \widehat J_t(\theta)=\widehat L_t(\theta)
                         +\frac{\|\theta-\theta_t\|_2^2}{2\eta_t}.
    \label{eq:proximal-solver}
\end{equation}
Thus the challenger seeks uncertainty, while the empirical solver minimizes
disagreement with the majority of $m$ old-solver responses. The theoretical
instance uses KL challenger regularization and a decreasing proximal step
size; it does not change the practical GRPO configuration.

Define
\begin{equation}
    \Phi_\beta(\theta)=\beta\log\E_{\pi_0}e^{[1-p_\theta(x)]/\beta},\qquad \kappa_T=\KL(q_T\|\pi_0).
    \label{eq:tilted-risk}
\end{equation}
Write $\Phi_*=\inf_\theta\Phi_\beta(\theta)$ and
$E_0=\Phi_\beta(\theta_0)-\Phi_*$. 

\paragraph{Initial answer separation.}
We assume that the initial solver satisfies, $\pi_0$-almost surely,
\begin{equation}
    P_{\theta_0}(a_0(x)\mid x)
       -\max_{a\ne a_0(x)}P_{\theta_0}(a\mid x)\geq\gamma_0>0.
    \label{eq:initial-answer-margin}
\end{equation}
Our local result derives later-round separation by bounding total
parameter movement, under the initialization and error-budget condition below.

\begin{theorem}[Worst-case uncertainty bound]
\label{thm:optimal-challenger}
Run the idealized variant of Algorithm~\ref{algorithm:mcmc-deo} in which
LLM-MH is replaced by exact sampling, with
\eqref{eq:concentration-rewards}--\eqref{eq:proximal-solver},
$n_t$ conditionally iid tasks per round from $q_t$ at each iteration,
$m$ fresh labeling responses per task, and exact empirical minimization
($\varepsilon_{\mathrm{opt},t}=0$). Assume at most $K\geq2$ canonical
answers, \eqref{eq:initial-answer-margin}, and globally $L$-smooth
fixed-answer losses with gradient norm at most $G>0$.
Assume conditional loss-class Rademacher complexity at most
$C_{\mathcal F}/\sqrt{n_t}$ and the local PL inequality
$\|\nabla\Phi_\beta(\theta)\|^2\geq2\mu[\Phi_\beta(\theta)-\Phi_*]$
on $\|\theta-\theta_0\|\leq r_0:=\gamma_0/(4G)$.
Fix $\delta\in(0,1)$ and use $\eta_t=18/[\mu(t+s)]$, with $s$ and the initialization/sampling
conditions specified in Appendix~\ref{app:initial-localization}.
With probability at least $1-\delta$, $\Phi_\beta(\theta_T)\leq B_T$ for
\begin{equation}
    B_T:=\Phi_*
      +\underbrace{\mathcal O\!\left(\frac{E_0}{T^2}\right)}_{\text{decaying optimization error}}
      +\underbrace{\widetilde{\mathcal O}\!\left(
          \frac1{T^2}\sum_{t=0}^{T-1}\frac{(t+1)^2C_{\cF}}{\sqrt{n_t}}
          +TK e^{-m\gamma_0^2/8}\right)}_{\text{statistical error}}.
    \label{eq:explicit-statistical-bound}
\end{equation}
The notation hides constants and confidence
logarithms. 
Consequently,
\begin{equation}
\boxed{
    \sup_{D:\,\KL(D\|\pi_0)\leq\kappa_T}
       \E_D[1-p_{\theta_T}(x)]
       \leq b_T:=\frac{1-e^{-\min\{B_T,1\}/\beta}}{1-e^{-1/\beta}}.
}
    \label{eq:induced-finite-sample-bound}
\end{equation}
\end{theorem}
The exact bound, including inexact-optimization errors, is
given in \eqref{eq:exact-statistical-bound}.
\paragraph{DEO learns distributionally robust reasoning.}
The bound controls the final solver's answer uncertainty over the induced
KL neighborhood of the base distribution $\pi_0$. The bound decreases in terms of the number of generated tasks and rollout
budgets.
The residual $\Phi_*$ accounts for the model class's irreducible uncertainty.

\paragraph{What drives the bound?}
At the current solver, disagreement with its old modal answer touches and
upper-bounds uncertainty. Therefore
\begin{equation}
    \E_{q_t}\left.\nabla_\theta r_s(x,\theta;a_t(x))\right|_{\theta=\theta_t}
       =\nabla\Phi_\beta(\theta_t).
\end{equation}
The advantage of sampling from the current challenger is that it follows the tilted uncertainty
potential, whereas base-law sampling (sampling from base task distribution $\pi_0$) generally follows mean uncertainty. This is why DEO can yield distributionally robust guarantee. The proof is deferred to Appendix~\ref{app:proof}.

An initial modal-answer correctness assumption further converts this
concentration guarantee into a bound on sampled-answer error.
\begin{corollary}[From concentration to answer accuracy]
\label{cor:accuracy}
Under Theorem~\ref{thm:optimal-challenger}, let $a^*(x)$ be the correct
answer, let $g(y)$ extract the answer of response $y$, and suppose
$\Pr_{\pi_0}[a_0(x)\ne a^*(x)]\leq\varepsilon_0$.
Set $\Delta_T=B_T-\Phi_*$ and
$b_*=(1-e^{-\Phi_*/\beta})/(1-e^{-1/\beta})$.
On the same probability-$1-\delta$ event,
\begin{equation}
\boxed{
\begin{aligned}
    &\sup_{D:\,\KL(D\|\pi_0)\leq\kappa_T}
       \Pr_{x\sim D,\,y\sim\pi_{\theta_T}(\cdot\mid x)}[g(y)\neq a^*(x)] \lesssim \min\!\left\{1,\;
       \varepsilon_0+b_*+\sqrt{\frac{b_*}{2\beta}}
       +\Delta_T+\sqrt{\Delta_T}
       \right\}.
\end{aligned}
}
    \label{eq:robust-accuracy-bound}
\end{equation}
The explicit remainder is given in \eqref{eq:accuracy-explicit-remainder}.
If $\Phi_*=0$, the bound is $\varepsilon_0+O_\beta(T^{-1/2})$,
capped at one. If $\varepsilon_0=0$, the sharper bound is
$b_T\leq b_*+O_\beta(\Delta_T)$.
\end{corollary}

Because the modal answer is preserved, the limiting error may exceed
the initial modal-answer error $\varepsilon_0$: the guarantee describes
robust sharpening of single-sample accuracy toward the initial
majority-vote accuracy, not the acquisition of answers the initial model
could not produce. The non-vanishing terms are an error certificate, not the
minimum robust error over the model class, and hold over the induced radius
$\kappa_T$ rather than a fixed neighborhood. Proofs appear in
Appendices~\ref{app:proof}--\ref{app:accuracy}; approximate sampling is
discussed in Appendix~\ref{app:sampler-error}.


\section{Experiments}
\label{sec:experiments}

We instantiate DEO for \emph{mathematical reasoning} and organize the
evaluation around four questions: whether challenger training is necessary
for competitive performance, how the
methods compare in computational cost, whether task refinement adds value and whether DEO can use an API-only
generator. Implementation details are in Appendix~\ref{app:practical-details}.

\paragraph{Models and comparisons.}
We evaluate Qwen3-4B-Base, Qwen3-8B-Base, and
OctoThinker-3B-Hybrid-Base through three self-evolving rounds, following the setup in~\cite{huang2025r,xia2025agent0}.  We compare with
R-Zero~\citep{huang2025r}, which trains a challenger, and \emph{no walk},
which samples tasks independently from the same frozen base model without LLM based Metropolis walk.
DEO and no walk each start from 2000 candidates per round and share the
labeling, filtering, and nominal solver-update configuration
(Appendix~\ref{app:experimental-protocol}).
\begin{figure}[t]
\centering
\begin{minipage}{0.49\linewidth}
\centering
\begin{tikzpicture}
\begin{axis}[
    width=\linewidth, height=4.1cm,
    ybar, bar width=12pt,
    symbolic x coords={Octo-3B, Qwen-4B,Qwen-8B}, xtick=data,
    enlarge x limits=0.45,
    ymin=0, ymax=62, ytick={0,15,30,45,60},
    ylabel={Hours}, title={(a) Wall-clock time},
    axis x line*=bottom, axis y line*=left,
    axis line style={black!50}, tick style={black!50},
    ymajorgrids, grid style={black!10},
    tick label style={font=\footnotesize},
    label style={font=\footnotesize},
    title style={font=\small},
    nodes near coords,
    nodes near coords style={font=\scriptsize, /pgf/number format/fixed,
      /pgf/number format/precision=1, /pgf/number format/fixed zerofill},
]
\addplot[fill=DEORuntimeRed, draw=DEORuntimeRed] coordinates {(Octo-3B,29.4) (Qwen-4B,30.5) (Qwen-8B,52.1)};
\addplot[fill=DEORuntimeGreen, draw=DEORuntimeGreen] coordinates {(Octo-3B,12.6) (Qwen-4B,14.6) (Qwen-8B,20.6)};
\end{axis}
\end{tikzpicture}
\end{minipage}\hfill
\begin{minipage}{0.49\linewidth}
\centering
\begin{tikzpicture}
\begin{axis}[
    width=\linewidth, height=4.1cm,
    ybar, bar width=12pt,
    symbolic x coords={Octo-3B, Qwen-4B, Qwen-8B}, xtick=data,
    enlarge x limits=0.45,
    ymin=0, ymax=65, ytick={0,10,20,30,40,50,60},
    ylabel={Accuracy (\%)}, title={(b) Math AVG7 Accuracy},
    axis x line*=bottom, axis y line*=left,
    axis line style={black!50}, tick style={black!50},
    ymajorgrids, grid style={black!10},
    tick label style={font=\footnotesize},
    label style={font=\footnotesize},
    title style={font=\small},
    nodes near coords,
    nodes near coords style={font=\scriptsize, /pgf/number format/fixed,
      /pgf/number format/precision=2, /pgf/number format/fixed zerofill},
]
\addplot[fill=DEORuntimeRed, draw=DEORuntimeRed] coordinates {(Octo-3B, 25.94) (Qwen-4B,45.93) (Qwen-8B,52.88)};
\addplot[fill=DEORuntimeGreen, draw=DEORuntimeGreen,
    nodes near coords style={yshift=7pt}] coordinates {(Octo-3B, 28.30) (Qwen-4B,47.57) (Qwen-8B,53.60)};
\end{axis}
\end{tikzpicture}
\end{minipage}
\par\smallskip
{\small
  \textcolor{DEORuntimeRed}{\rule{9pt}{6pt}}\; R-Zero
  \hspace{1.5em}
  \textcolor{DEORuntimeGreen}{\rule{9pt}{6pt}}\; DEO}
\caption{Wall-clock time and iteration-three AVG7 accuracy of DEO and R-Zero.
DEO uses 57.1\%, 52.1\%, and 60.5\% less wall-clock time on OctoThinker-3B,
Qwen3-4B, and Qwen3-8B, respectively, while achieving competitive accuracy.
The Qwen R-Zero runs use larger native question pools
(Section~\ref{sec:challenger-comparison}).}
\label{fig:reported-runtime}
\end{figure}
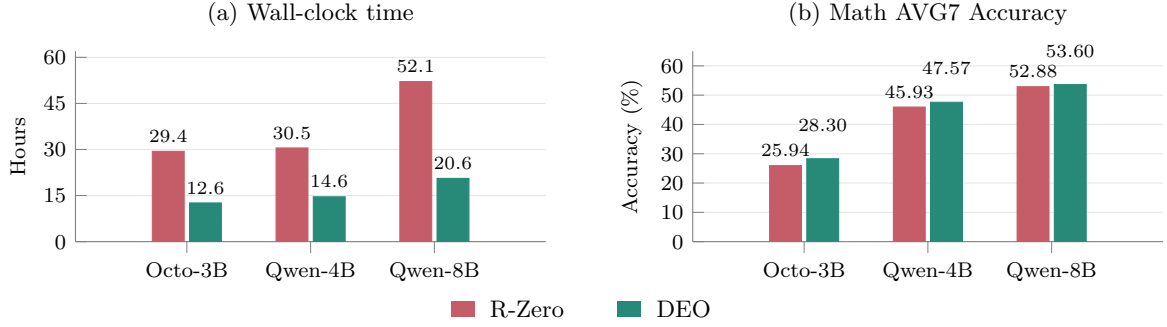

\paragraph{Evaluation.}
The primary comparison reports iteration-three accuracy on
MATH-500~\citep{hendrycks2021math,lightman2023lets},
GSM8K~\citep{cobbe2021training}, AMC~\citep{maaAMC},
Minerva~\citep{lewkowycz2022solving},
OlympiadBench~\citep{he2024olympiadbench}, and the 2024 and 2025
editions of AIME~\citep{maaAIME}.
AVG7 is their unweighted mean; HARD5 averages the last five benchmarks.
All methods use the same archived boxed-answer grading protocol.
Additional Omni-MATH~\citep{gao2024omnimath} results are reported in
Appendix~\ref{app:omni-math}.
We additionally measure transfer to MMLU-Pro~\citep{wang2024mmlupro},
SuperGPQA~\citep{mapteam2025supergpqa}, and BBEH~\citep{kazemi2025bbeh}
at all three scales, with their mean denoted AVG3. Table entries are
percentages. 
The evaluation protocol, repeated-question handling, and
checkpoint conventions are detailed in Appendix~\ref{app:experimental-protocol}.

\subsection{DEO competes without training a challenger, using half the training time}
\label{sec:challenger-comparison}

Table~\ref{tab:main-results} compares each pretrained checkpoint with
the three methods after the same number of self-evolving rounds.
DEO reaches AVG7 scores of 47.57 at 4B and 53.60 at 8B, exceeding R-Zero
by 1.64 and 0.72 percentage points. The corresponding HARD5 gains are
1.80 and 0.99 points. On OctoThinker-3B, DEO reaches 28.30 AVG7 versus
25.94 for R-Zero, with MATH-500 improving from 50.20 to 52.60.
Thus competitive aggregate performance does not require updating the
challenger's parameters in these experiments. The advantage is not uniform
over rounds: at 4B, R-Zero leads in iterations one and two
(Table~\ref{tab:iteration-results}).
On the same eight-H100 node, DEO reduces wall-clock time by 52--61\%
(Figure~\ref{fig:reported-runtime}). 
No walk takes 5.9, 4.9, and 6.2 hours through iteration three on
OctoThinker-3B, Qwen3-4B, and Qwen3-8B, respectively. 

\paragraph{General-domain transfer.}
At 4B, DEO
improves AVG3 from 30.13 to 32.17, above no walk's 31.89 but below
R-Zero's 32.59. At 8B, DEO attains the highest AVG3, 35.04, versus
34.62 for no walk and 34.30 for R-Zero. DEO has the highest BBEH score
at both Qwen scales, while R-Zero leads MMLU-Pro and, on OctoThinker-3B,
AVG3. Relative transfer gains thus depend on the model and benchmark.

\begin{table} 
\centering
\small
\setlength{\tabcolsep}{3.5pt}
\caption{Base checkpoints and iteration-three results (accuracy, \%).
DEO and no walk use 2000 initial candidates; the Qwen R-Zero runs retain
their native larger pools. Implementation settings are in
Appendix~\ref{app:practical-details}. Base denotes the pretrained checkpoint
before self-evolution. Bold marks the best score in each
column within each base-model block, including ties.}
\label{tab:main-results}
\resizebox{ \linewidth}{!}{\begin{tabular}{@{}lrrrrrrrrrrrrr@{}}
\toprule
Method & \shortstack{MATH\\500} & GSM8K & AMC & Minerva
& \shortstack{Olympiad\\Bench} & \shortstack{AIME\\2024}
& \shortstack{AIME\\2025} & AVG7 & HARD5 & MMLU-Pro & SuperGPQA & BBEH & AVG3 \\
\midrule
\multicolumn{14}{@{}l}{\textit{OctoThinker-3B-Hybrid-Base}} \\
\addlinespace[2pt]
Base & 44.60 & 61.49 & 17.42 & 19.12 & 14.67 & \textbf{6.67} & {0.00} & 23.42 & 11.58 & 8.20 &  5.33 & 0.97 & 4.83 \\
No walk & 48.40 & 73.24 & 25.00 & \textbf{27.21} & \textbf{18.67} & 0.00 & {0.00} & 27.50 & \textbf{14.18} & 17.73 & 9.70 &3.30 & 10.24 \\
R-Zero & 50.20 & 73.09 & 12.50 & \textbf{27.21} & 15.26 & 3.33 & {0.00} & 25.94 & 11.66 & \textbf{19.10} & 10.47 & \textbf{4.00} & \textbf{11.19} \\
DEO & \textbf{52.60} & \textbf{74.75} & \textbf{25.08} & 26.47 & 15.85 & 3.33 & {0.00} & \textbf{28.30} & 14.15 &17.97 & \textbf{10.83} & 3.70 & 10.83 \\
\midrule
\multicolumn{14}{@{}l}{\textit{Qwen3-4B-Base}} \\
\addlinespace[2pt]
Base & 72.60 & 81.90 & 48.40 & 36.80 & 39.10 & \textbf{10.00} & \textbf{16.60} & 43.63 & 30.18 & 55.13 & 27.07 & 8.20 & 30.13 \\
No walk & 75.80 & 91.51 & 52.58 & 47.43 & 39.11 & \textbf{10.00} & 3.33 & 45.68 & 30.49 & 58.73 & 28.67 & 8.27 & 31.89 \\
R-Zero & 74.00 & \textbf{91.70} & 47.30 & \textbf{51.80} & 40.40 & 9.60 & 6.70 & 45.93 & 31.16 & \textbf{59.37} & \textbf{29.53} & 8.87 & \textbf{32.59} \\
DEO & \textbf{76.80} & 91.43 & \textbf{57.34} & 45.96 & \textbf{41.48} & 6.67 & 13.33 & \textbf{47.57} & \textbf{32.96} & 58.10 & 28.50 & \textbf{9.90} & 32.17 \\
\midrule
\multicolumn{14}{@{}l}{\textit{Qwen3-8B-Base}} \\
\addlinespace[2pt]
Base & 70.00 & 88.93 & 52.50 & 43.38 & 40.74 & 10.00 & 3.33 & 44.13 & 29.99 & 60.23 & 31.33 & 9.87 & 33.81 \\
No walk & 81.20 & 93.78 & \textbf{64.53} & \textbf{54.41} & 45.19 & 10.00 & 13.33 & 51.78 & 37.49 & 62.17 & \textbf{32.17} & 9.53 & 34.62 \\
R-Zero & \textbf{81.40} & 93.86 & 62.81 & 53.68 & 45.04 & 16.67 & \textbf{16.67} & 52.88 & 38.97 & \textbf{62.83} & 31.63 & 8.43 & 34.30 \\
DEO & 81.20 & \textbf{94.16} & 62.81 & \textbf{54.41} & \textbf{45.93} & \textbf{20.00} & \textbf{16.67} & \textbf{53.60} & \textbf{39.96} & 62.23 & \textbf{32.17} & \textbf{10.73} & \textbf{35.04} \\

\bottomrule
\end{tabular}}
\end{table}

\subsection{LLM mutation refines the question pool}
\label{sec:in-band}

\begin{figure}  [t]
\centering
\includegraphics[width= \linewidth]{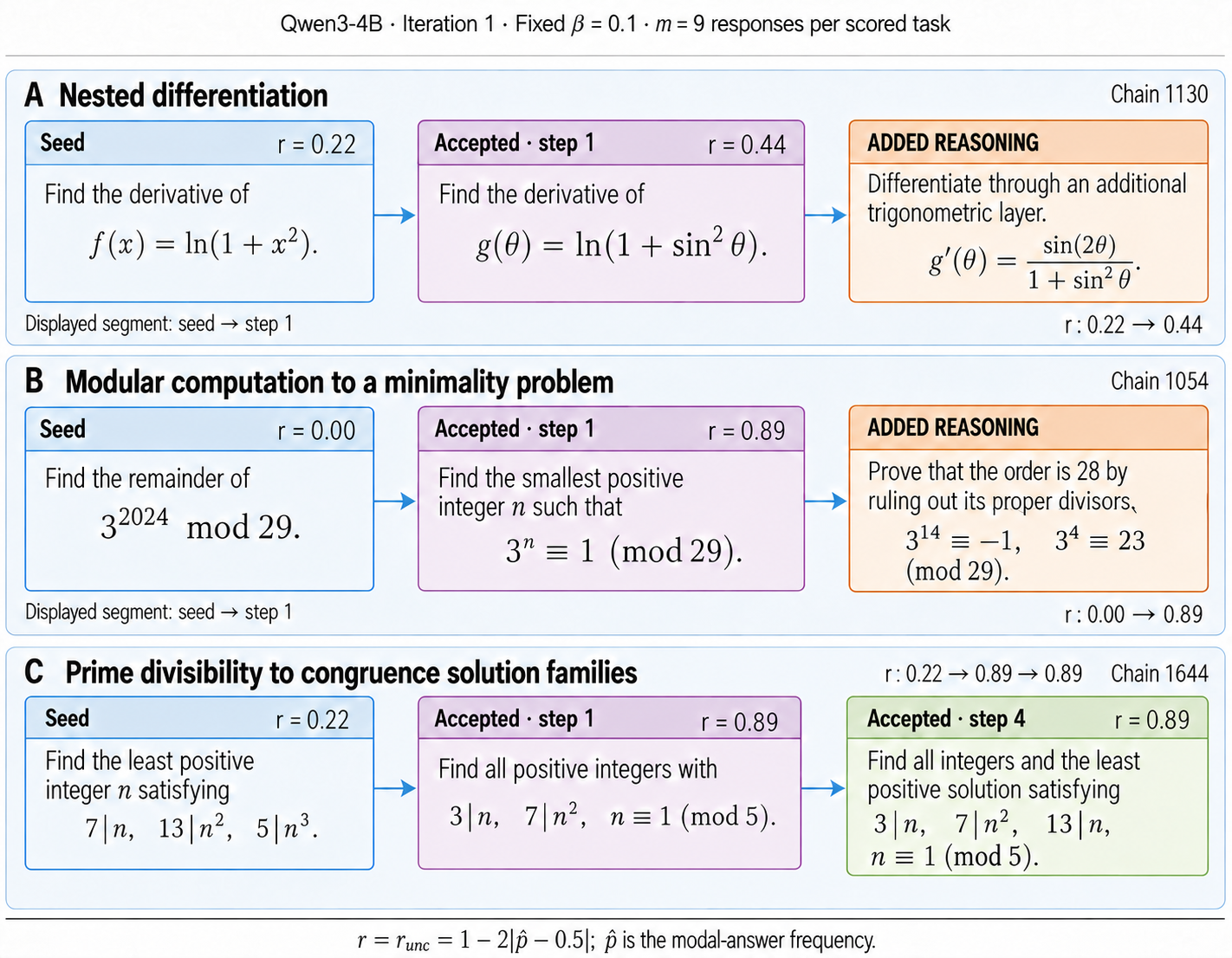}
\caption{Accepted LLM mutations add reasoning requirements (Qwen3-4B,
iteration one). A and B show the seed and first accepted mutation, with the
added reasoning explained on the right; C shows both accepted mutations.
Each shown mutation raises the uncertainty score and is accepted with
probability one.}
\label{fig:walk-examples}
\end{figure}

The no-walk comparison holds the candidate-pool and solver-update
configurations fixed. In Table~\ref{tab:main-results}, DEO improves AVG7
over no walk by 1.89 points at 4B, 1.82 points at 8B, and 0.79 points
on OctoThinker-3B. The Qwen HARD5 gains are 2.47 points at both scales,
while no walk retains a slight OctoThinker-3B HARD5 edge (14.18 versus
14.15). The robustness guarantee of Section~\ref{sec:theory} does not
imply dominance on every benchmark. The walk adds generation and solver
scoring (up to $L_{\mathrm w}=5$ proposals per task), so this comparison
is matched in pool size and solver-update configuration, not total compute.
\paragraph{The trajectory of LLM mutation}
Figure~\ref{fig:walk-examples} illustrates three archived trajectories from
an instrumented Qwen3-4B run with the same fixed-temperature recipe.
The frozen LLM introduces an additional trigonometric composition in a
derivative problem (A), changes modular exponentiation into a
multiplicative-order problem requiring a minimality argument (B), and
replaces prime-divisibility conditions with congruence solution families
before adding another constraint (C). The recorded uncertainty scores
increase from $0.22$ to $0.44$, $0$ to $0.89$, and $0.22$ to $0.89$,
respectively; the last accepted transition in C leaves the score at
$0.89$. These selected examples show how refinement can add reasoning
requirements while increasing the solver's measured uncertainty.

\begin{wraptable}{r}{0.5 \textwidth}
\vspace{-1\baselineskip}
\centering
\footnotesize
\setlength{\tabcolsep}{3pt}
\caption{Question counts for Qwen3-4B, iterations 1--3. In-band counts
precede the pseudo-label and format checks; retained counts follow them.
DEO and no walk each start with 2000 candidates.}
\label{tab:in-band}
\resizebox{\linewidth}{!}{%
\begin{tabular}{@{}lrrrrrr@{}}
\toprule
& \multicolumn{3}{c}{In-band}
& \multicolumn{3}{c}{Retained} \\
\cmidrule(lr){2-4}\cmidrule(l){5-7}
Method & Iter 1 & Iter 2 & Iter 3 & Iter 1 & Iter 2 & Iter 3 \\
\midrule
No walk & 1029 & 970 & 1044 & 977 & 913 & 1002 \\
DEO & 1386 & 1358 & 1393 & 1246 & 1208 & 1228 \\

\bottomrule
\end{tabular}}
\end{wraptable}

\paragraph{LLM mutation increases in-band questions.}
Table~\ref{tab:in-band} reports the archived Qwen3-4B counts under the
training agreement-band criterion defined in Appendix~\ref{app:practical-filter}. In short, in-band counts how many generated tasks fall in the target uncertainty range (the ``sweet spot''), excluding extremely hard and extremely easy questions.
Pooling rounds one through three, the in-band fraction is 50.72\% for
no walk and 68.95\% for DEO, with 357, 388, and 349 additional in-band candidates,
respectively. The advantage remains after filtering: 3682 retained tasks
in total versus 2892 for no walk. This shows that solver-guided refinement
increases the operational yield of the candidate pool with a frozen
generator. These counts use the same cached responses that drove
selection, so they measure yield under the walk's own score, not
independently verified difficulty or label correctness.

\subsection{DEO works with an API-only generator}
\label{sec:claude-generator}
Removing challenger training permits a task generator whose parameters
are inaccessible. We replace the local Qwen3-4B generator with Claude
Haiku 4.5 for seed generation and mutation. The Qwen3-4B solver still
supplies the scores and pseudo-labels; Claude's answers are never training
targets. The task budget and training recipe are unchanged
(Appendix~\ref{app:experimental-protocol}).
At iteration three, the Claude-generator run reaches 49.60 AVG7 and
35.77 HARD5, compared with 47.57 AVG7 and 32.96 HARD5 for the
local-generator DEO run. Gains are concentrated in Minerva, OlympiadBench, and
AIME 2024; MATH-500 is unchanged and AMC decreases. Thus a frozen API-only
model can improve the task distribution used to train a local solver. 
\begin{table}  
\centering
\small
\setlength{\tabcolsep}{3.5pt}
\caption{DEO with a Qwen3-4B versus Claude Haiku 4.5 generator/mutator;
Qwen3-4B solver, iteration three (\%). Bold: better score, including ties. For the Claude one, the pseudo-labels for training were still produced by last-round solver $\theta_t$. The Claude generator+mutator can notably boost the performance, while the R-Zero type of method cannot leverage that.}
\label{tab:claude-generator}
\resizebox{ \linewidth}{!}{%
\begin{tabular}{@{}lrrrrrrrrr@{}}
\toprule
Generator & \shortstack{MATH\\500} & GSM8K & AMC & Minerva
& \shortstack{Olympiad\\Bench} & \shortstack{AIME\\2024}
& \shortstack{AIME\\2025} & AVG7 & HARD5 \\
\midrule
Qwen3-4B & \textbf{76.80} & 91.43 & \textbf{57.34} & 45.96 & 41.48
& 6.67 & \textbf{13.33} & 47.57 & 32.96 \\
Claude & \textbf{76.80} & \textbf{91.51} & 52.50 & \textbf{50.37}
& \textbf{42.67} & \textbf{20.00} & \textbf{13.33}
& \textbf{49.60} & \textbf{35.77} \\
\bottomrule
\end{tabular}}
\end{table}

\section{Conclusion}

DEO uses the challenger objective to define a sampling target and
approximates it with LLM mutation and Metropolis selection, training only
the solver. Idealized DEO provably learns distributionally robust
reasoning; in practice DEO matches challenger training at about half the
wall-clock time, and can use an API-only generator.
\emph{Limitations:} The guarantees cover idealized DEO only, while in practical algorithm we use LLM based MH walk to simulate the sampling process. Hence there is a gap between theory and the algorithm we actually use. In addition, the walk operates per question within a round and has no cross-iteration
memory, whereas R-Zero-style challengers evolve continuously across
iterations. Future work includes memory mechanisms for the walk and applying
DEO to agent training. (See more discussion in Appendix~\ref{app:limitations}).

\IfFileExists{references.bib}{%
    \bibliographystyle{plainnat}
    \bibliography{references}
}{%
    
}
\clearpage
\appendix
\startcontents[appendix]
\section*{Appendix}
\section*{Contents}
\printcontents[appendix]{}{1}{\setcounter{tocdepth}{2}}

\section{Limitations}
\label{app:limitations}
\paragraph{Theory versus implementation.}
Theorem~\ref{thm:optimal-challenger} and Corollary~\ref{cor:accuracy} cover
idealized DEO: exact iid sampling from $q_t$, the monotone reward
$1-p_\theta$, fresh labeling responses, and a proximal L2 solver update. The
implemented walk omits the base-density/proposal ratio, uses noisy scores and
finitely many sweeps, uses a triangular score with a repetition penalty and
an agreement filter, reuses walk responses as labels, and trains with GRPO.
Appendix~\ref{app:sampler-error} shows how a sampler error $d_t$ would enter
the bound, but we do not estimate $d_t$ for the LLM walk. The theorem also
relies on local PL and a localization budget that we cannot verify for LLMs.

\paragraph{What the guarantee certifies.}
Supervision comes from the solver's own majority vote, and the proof keeps the
initial modal answer fixed. The accuracy bound therefore cannot fall below the
initial modal-answer error $\varepsilon_0$. It describes robust sharpening of
single-sample accuracy toward majority-vote accuracy, not the discovery of
answers the initial model could not produce. The robustness radius
$\kappa_T$ is induced by the final model rather than fixed in advance.

\paragraph{Label and task quality.}
Pseudo-labels are unverified majority answers. Higher solver uncertainty can
reflect ill-posed or ambiguous tasks as well as harder ones. The in-band
statistics in Section~\ref{sec:in-band} reuse the responses that drove
selection, so they are not an independent measure of difficulty or
correctness.


\paragraph{Design.}
The walk operates per question within a round and has no cross-iteration
memory, whereas R-Zero-style challengers evolve continuously across
iterations. Future work includes memory mechanisms for the walk and applying
DEO to agent training.

\section{Proof of Theorem~\ref{thm:optimal-challenger}}
\label{app:proof}

\subsection{Definitions and objective compatibility}
\label{app:theory-definitions}

Let $g$ extract a valid canonical answer, with a fixed tie-breaking rule.
The answer law and the empirical plurality rule used in the analysis are
\begin{align}
    P_\theta(a\mid x)&=\Pr_{y\sim\pi_\theta(\cdot\mid x)}[g(y)=a],\qquad
    a_t(x)\in\arg\max_aP_{\theta_t}(a\mid x),
    \label{eq:old-majority-definition}\\
    \widehat z_{t,i}&=\mathcal A_m(y_{t,i,1:m})\in
       \arg\max_a\sum_{j=1}^m\ind\{g(y_{t,i,j})=a\}.
    \label{eq:empirical-concentration-label}
\end{align}
Labels are fixed when optimizing the candidate solver. Its loss against
the old population mode is
\begin{equation}
    r_s(x,\theta;a_t(x))=1-P_\theta(a_t(x)\mid x),\qquad
    L_t(\theta)=\E_{x\sim q_t}r_s(x,\theta;a_t(x)).
    \label{eq:modal-disagreement-loss}
\end{equation}
The empirical L2 instance sets
\begin{equation}
    \mathcal R_s(\pi_\theta,\pi_{\theta_t})
       =\tfrac12\|\theta-\theta_t\|_2^2,\qquad\tau_t=\eta_t^{-1},
    \label{eq:l2-instance}
\end{equation}
in a fixed parameterization. This is separate from the KL regularizer
on the challenger's question law.

The KL-optimal challenger has the explicit exponential-tilt form
\begin{equation}
    q_\beta^\theta(x)=\frac{\pi_0(x)e^{[1-p_\theta(x)]/\beta}}{Z_\theta},
    \qquad Z_\theta=\E_{\pi_0}e^{[1-p_\theta(x)]/\beta},
    \quad q_t=q_\beta^{\theta_t}.
    \label{eq:ideal-question-target}
\end{equation}

For later use, define worst-case uncertainty by
\begin{equation}
    \operatorname{Unc}_{\kappa}(\theta)
       :=\sup_{D:\,\KL(D\|\pi_0)\leq\kappa}
               \E_{x\sim D}[1-p_\theta(x)].
    \label{eq:worst-case-concentration}
\end{equation}
The exact tilt induces the radius and achieves the risk
\begin{align}
    \kappa_\beta(\theta)
       &=\frac{\E_{q_\beta^\theta}[1-p_\theta(x)]-\Phi_\beta(\theta)}{\beta},
    \label{eq:natural-radius}\\
    \operatorname{Unc}_{\kappa_\beta(\theta)}(\theta)
       &=\E_{q_\beta^\theta}[1-p_\theta(x)]
        =\Phi_\beta(\theta)+\beta\kappa_\beta(\theta).
    \label{eq:natural-adversarial-risk}
\end{align}
We prove these identities below. The ball is induced by the final model;
it need not be fixed over training, and $q_\beta^{\theta_T}$ need not equal
the last training distribution $q_{T-1}$.

All iterates in this proof are the empirical iterates of the idealized
(exact-sampling) variant of Algorithm~\ref{algorithm:mcmc-deo}. Write $\Phi=\Phi_\beta$,
$c_\theta(x)=r_c(x,\theta)$ and $g_t=\nabla\Phi(\theta_t)$. The losses
evaluated against population majority labels and their proximal minimizers
below are comparison quantities in the proof, not additional
updates performed by the algorithm.

Since $\max_aP_\theta(a\mid x)\geq P_\theta(a_t(x)\mid x)$, the
definitions in \eqref{eq:concentration-rewards} give
\begin{align}
    r_c(x,\theta)&\leq r_s(x,\theta;a_t(x))
                                  &&\text{for every $\theta$},\nonumber\\
    r_c(x,\theta_t)&=r_s(x,\theta_t;a_t(x)),\nonumber\\
    \nabla_\theta r_c(x,\theta_t)
       &=\left.\nabla_\theta r_s(x,\theta;a_t(x))\right|_{\theta=\theta_t}.
    \label{eq:reward-compatibility}
\end{align}
The last equality holds when the old modal answer is locally unique and
the answer probabilities are differentiable. Thus the touching upper
bound and gradient alignment follow from the specified objectives.
The upper bound is valid even outside the initial neighborhood; later
we prove that the actual iterates preserve the initial modal answer.

\paragraph{Regularity.}
Both $r_s$ and $r_c$ lie in $[0,1]$. Assume that each fixed-answer loss
$r_s(x,\cdot;z)$ has an $L$-Lipschitz gradient of norm at most $G$,
uniformly in $x,z$. It follows that $r_c(x,\cdot)$ is $G$-Lipschitz;
on the initial-margin neighborhood below it is also differentiable.
All question-wise conditions hold $\pi_0$-almost surely. Work in
$\mathbb R^d$ with $\beta>0$ fixed and $L+G^2/\beta>0$.

The current optimal challenger therefore gives
\begin{equation}
\begin{aligned}
    \nabla L_t(\theta_t)
        &=\E_{q_t}\left[\left.\nabla_\theta r_s(x,\theta;a_t(x))
                              \right|_{\theta=\theta_t}\right]\\
        &=\E_{q_t}\nabla_\theta r_c(x,\theta_t)
         =\nabla\Phi_\beta(\theta_t).
\end{aligned}
    \label{eq:gradient-alignment}
\end{equation}
Sampling from $q_t$ aligns the expected solver-update gradient with the
exponentially tilted uncertainty potential. Sampling from $\pi_0$
instead generally aligns only with mean uncertainty under the base law.

\subsection{Initial margin and the localization budget}
\label{app:initial-localization}

For completeness, we retain inexact empirical optimization in the proof;
the two-term display in Theorem~\ref{thm:optimal-challenger} specializes
to $\varepsilon_{\mathrm{opt},t}=0$. Fix $\delta\in(0,1)$ and define
\begin{equation}
\begin{aligned}
    \delta_m^0&:=\min\{1,(K-1)e^{-m\gamma_0^2/8}\},\\
    e_t&:=\frac{4C_{\mathcal F}}{\sqrt{n_t}}
       +2\sqrt{\frac{\log(2T/\delta)}{2n_t}}
       +2\delta_m^0+\varepsilon_{\mathrm{opt},t}.
\end{aligned}
    \label{eq:initial-margin-errors}
\end{equation}
Choose $H=L+G^2/\beta$, $s\geq\max\{3,36H/\mu\}$, and
$\eta_t=18/[\mu(t+s)]$. The sampling/optimization budget is
$e_t\leq\zeta/(t+s)^2$ for some $\zeta>0$.
Set $C_0=(s-1)E_0+3\zeta$ and $\Lambda_T=(T+s-2)(T+s-1)$.
The exact finite-sample bound is
\begin{equation}
\begin{aligned}
    B_T:={}&\Phi_*+\frac{(s-2)(s-1)E_0}{\Lambda_T}
       +\frac3{\Lambda_T}\sum_{t=0}^{T-1}(t+s-1)(t+s)e_t\\
       &\leq\Phi_*+\frac{C_0}{T+s-1}.
\end{aligned}
    \label{eq:exact-statistical-bound}
\end{equation}
For fixed $s$, the first excess term is $O(E_0/T^2)$.
The weighted finite-task errors are
$\widetilde O(T^{-2}\sum_{t<T}(t+1)^2/\sqrt{n_t})$, and the common
finite-rollout error contributes $O(TK e^{-m\gamma_0^2/8})$.
These give \eqref{eq:explicit-statistical-bound} for exact empirical ERM.
An inexact solver adds precisely
$3\Lambda_T^{-1}\sum_{t<T}(t+s-1)(t+s)\varepsilon_{\mathrm{opt},t}$;
it also consumes the same error and localization budgets.

Write $\Phi=\Phi_\beta$, $E_t=\Phi(\theta_t)-\Phi_*$, and
$\mathcal B_0=\{\theta:\|\theta-\theta_0\|\leq r_0\}$, where
$r_0=\gamma_0/(4G)$. The global fixed-answer gradient bound implies that,
for $\theta\in\mathcal B_0$ and any $a\ne a_0(x)$,
\begin{equation}
\begin{aligned}
    P_\theta(a_0(x)\mid x)-P_\theta(a\mid x)
       &\geq\gamma_0-2G\|\theta-\theta_0\|\\
       &\geq\gamma_0/2.
\end{aligned}
    \label{eq:initial-gap-preservation}
\end{equation}
This establishes unique, unchanged modal answers throughout a fixed
neighborhood without assuming anything about future iterates. Since the
tilt density relative to $\pi_0$ is strictly positive, the same statement
holds almost surely under every ideal challenger distribution.

Theorem~\ref{thm:optimal-challenger} assumes the PL inequality on
$\mathcal B_0$, not an a priori margin condition along the trajectory.
With $H=L+G^2/\beta$, $s\geq\max\{3,36H/\mu\}$,
$C_0=(s-1)E_0+3\zeta$, its explicit additional condition is
\begin{equation}
    \boxed{
    \frac{72\sqrt{2HC_0}}{\mu\sqrt{s-2}}
       +4\sqrt{\frac{18\zeta}{\mu(s-1)}}
       \leq \frac{\gamma_0}{4G}.}
    \label{eq:localization-budget}
\end{equation}
It requires sufficiently small initial excess potential and error budget
relative to the initial answer separation. It is a sufficient, conservative
local-basin condition, not a consequence of initial margin alone. Its
constants are independent of the horizon when $E_0,\zeta,s$ are fixed.
The empirical optimizer is still unconstrained in $\mathbb R^d$; no
projection, frozen-label training rule, or extra population update is added.
Labels are regenerated from the current solver at every round.

\paragraph{Finite sampling budgets.}
For a prescribed $T$, put
$A_{T,\delta}=2C_{\mathcal F}+\sqrt{\log(2T/\delta)/2}$.
For $\zeta>0$, the following sufficient choices ensure
$e_t\leq\zeta/(t+s)^2$:
\begin{equation}
\begin{aligned}
    n_t&\geq\frac{36A_{T,\delta}^2(t+s)^4}{\zeta^2},\\
    m&\geq\frac{8}{\gamma_0^2}
          \log_+\!\left(\frac{6(K-1)(T+s)^2}{\zeta}\right),\\
    \varepsilon_{\mathrm{opt},t}&\leq\frac{\zeta}{3(t+s)^2}.
\end{aligned}
    \label{eq:sufficient-finite-budgets}
\end{equation}
Here $\log_+(u)=\max\{0,\log u\}$, and positive integer sample sizes are
rounded upward. A common $n$ can use the largest displayed task budget.
The common rollout count $m$ remains fixed within a finite-horizon run;
it must increase with the horizon in an asymptotic family of runs.
These are sufficient theoretical budgets, not the settings used in the
experiments. The finite-$m$ term is an expected label-error contribution;
the proof does not require all sampled labels to be correct simultaneously.

\paragraph{Why decreasing steps alone are insufficient.}
The crude bound $\|\theta_{t+1}-\theta_t\|\leq G\eta_t$ for exact
empirical proximal minimization preserves a neighborhood if
$\sum_t\eta_t$ is small, but then it does not yield a vanishing
optimization term through the PL contraction. Our schedule has
$\sum_t\eta_t=\infty$. It is the decay of the update direction, together
with decreasing statistical errors, that makes the actual path length
finite. This is proved next rather than assumed.

\subsection{Uniform sampling error and finite-rollout labels}
\label{app:empirical}

Let $\mathcal F_{t,m}=\{(x,z)\mapsto r_s(x,\theta;z):\theta\in\mathbb R^d\}$.
Its sampling law is a fresh question from $q_t$ together with its
$m$-response pseudo-label. Conditional on the pre-round history, let
$\mathfrak R_{n_t}(\mathcal F_{t,m})$ denote its expected Rademacher
complexity. Under the stated capacity bound, use the deterministic envelope
\begin{equation}
    \varepsilon_{n,t}=2C_{\mathcal F}/\sqrt{n_t}
                   +\sqrt{\frac{\log(2T/\delta)}{2n_t}}.
    \label{eq:explicit-question-error}
\end{equation}
The theorem assumes $\mathfrak R_{n_t}(\mathcal F_{t,m})\leq
C_{\mathcal F}/\sqrt{n_t}$ uniformly over rounds and admissible histories.
This is a capacity assumption; bounded loss alone does not imply it.

For the following one-step estimates, condition on a history with
$\theta_t\in\mathcal B_0$. The frozen old solver and $q_t$
are then fixed. Each pair $(x_{t,i},\widehat z_{t,i})$ is sampled
independently from the law specified in Theorem~\ref{thm:optimal-challenger}. Let
\[
    \widetilde L_{t,m}(\theta)
       :=\E_{x\sim q_t,\,\widehat z\sim H_{t,m}(\cdot\mid x)}
                                      r_s(x,\theta;\widehat z),
\]
where $H_{t,m}$ is the law of the $m$-rollout pseudo-label. A conditional
uniform deviation bound for the bounded loss class gives
\[
    \sup_\theta|\widehat L_t(\theta)-\widetilde L_{t,m}(\theta)|
          \leq\varepsilon_{n,t}
\]
with conditional failure probability at most $\delta/T$. A union bound
yields a simultaneous event for all pre-exit rounds with probability
at least $1-\delta$. The stopping argument below proves that no exit
occurs. Independence across rounds is not required.

For the theoretical disagreement loss, let $a_t(x)$ denote the population modal
answer of the old solver. For any competing answer $a$, define the
$m$ independent variables
$W_j=\ind\{g(y_j)=a\}-\ind\{g(y_j)=a_t(x)\}$.
They lie in $[-1,1]$ and have mean at most $-\gamma_0/2$. A bounded-sum tail
inequality gives
$\Pr(\sum_jW_j\geq0)\leq e^{-m\gamma_0^2/8}$.
Taking a union bound over the $K-1$ competitors, including ties as possible
errors, gives
\begin{equation}
    \Pr[\widehat z\ne a_t(x)]
         \leq\delta_m^0:=\min\{1,(K-1)e^{-m\gamma_0^2/8}\}.
    \label{eq:finite-m-label-error}
\end{equation}
Since $r_s\in[0,1]$, replacing the population modal label by $\widehat z$
changes its expected loss by at most $\delta_m^0$, uniformly in $x,\theta$.
Consequently, with $L_t(\theta)=\E_{q_t}r_s(x,\theta;a_t(x))$,
\begin{equation}
    \sup_\theta|\widehat L_t(\theta)-L_t(\theta)|
             \leq\varepsilon_{n,t}+\delta_m^0=:b_t.
    \label{eq:empirical-loss-comparison}
\end{equation}
This controls disagreement with the old solver's modal answer, without
assuming that this answer is correct.

\subsection{Relating empirical ERM to a comparison function}
\label{app:inexact}

Define the comparison function and its minimizer
\begin{equation}
    F_t(\theta):=L_t(\theta)
                     +\frac{\|\theta-\theta_t\|_2^2}{2\eta_t},
    \qquad z_t:=\arg\min_\theta F_t(\theta).
    \label{eq:comparison-proximal-function}
\end{equation}
The loss $L_t$ is $L$-smooth and hence $L$-weakly convex. Thus $F_t$ is
$M$-strongly convex with $M=\eta_t^{-1}-L\geq1/(2\eta_t)>0$.
The bounded loss and coercive quadratic imply that $z_t$ exists and is
unique. The empirical algorithm does not compute $z_t$.

The same quadratic appears in $F_t$ and $\widehat J_t$. Comparing the actual
empirical optimizer to $z_t$ on the simultaneous deviation event gives
\begin{align*}
    F_t(\theta_{t+1})
      &\leq\widehat J_t(\theta_{t+1})+b_t\\
      &\leq\widehat J_t(z_t)+\varepsilon_{\mathrm{opt},t}+b_t\\
      &\leq F_t(z_t)+2b_t+\varepsilon_{\mathrm{opt},t}.
\end{align*}
Therefore
\begin{equation}
    F_t(\theta_{t+1})-F_t(z_t)
       \leq e_t:=2\varepsilon_{n,t}+2\delta_m^0
                                  +\varepsilon_{\mathrm{opt},t}.
    \label{eq:empirical-oracle-error}
\end{equation}
Strong convexity also yields
$\|\theta_{t+1}-z_t\|_2^2\leq2e_t/M$.

\subsection{Descent along the empirical iterates}

For $q_\beta^\theta=\pi_0e^{c_\theta/\beta}/Z_\theta$, direct substitution
gives the Gibbs identity
\begin{equation}
    \E_Dc_\theta-\beta\KL(D\|\pi_0)
       =\Phi(\theta)-\beta\KL(D\|q_\beta^\theta).
    \label{eq:variational-identity-proof}
\end{equation}
Boundedness and the gradient envelope justify differentiation under the
expectation at each visited iterate. The derived identity
\eqref{eq:reward-compatibility} gives
\[
    g_t=\E_{q_t}\nabla c_{\theta_t}
       =\E_{q_t}\left[\left.\nabla_\theta r_s(x,\theta;a_t(x))
                              \right|_{\theta=\theta_t}\right]
       =\nabla L_t(\theta_t).
\]
For any candidate $\theta'=\theta_t+\Delta$, the tilt also gives
\[
    \Phi(\theta')-\Phi(\theta_t)
       =\beta\log\E_{q_t}
                     e^{[c_{\theta'}(x)-c_{\theta_t}(x)]/\beta}.
\]
Since $c_{\theta'}-c_{\theta_t}\in[-G\|\Delta\|_2,G\|\Delta\|_2]$,
the inequality
$\log\E e^{s(W-\E W)}\leq s^2(b-a)^2/8$ for $W\in[a,b]$ implies
\begin{align}
    \Phi(\theta')-\Phi(\theta_t)
       &\leq\E_{q_t}[c_{\theta'}-c_{\theta_t}]
                                  +\frac{G^2}{2\beta}\|\Delta\|_2^2\nonumber\\
       &\leq L_t(\theta')-L_t(\theta_t)
                                  +\frac{G^2}{2\beta}\|\Delta\|_2^2.
    \label{eq:descent-upper-model}
\end{align}
The touching upper bound justifies the second line even when the modal
answer changes between iterates.

Write $\Delta_t=\theta_{t+1}-\theta_t$ and
$\Delta_t^*=z_t-\theta_t$, and set
\[
    A:=\frac1{2\eta_t}-\frac{G^2}{2\beta},\qquad
    B:=\frac1{\eta_t}+L,\qquad M:=\frac1{\eta_t}-L.
\]
The step-size condition gives $A\geq1/(4\eta_t)$,
$B\leq3/(2\eta_t)$ and $M\geq1/(2\eta_t)$. Since
$F_t(\theta_{t+1})\leq F_t(\theta_t)+e_t$,
\eqref{eq:descent-upper-model} yields
\[
    \Phi(\theta_{t+1})\leq\Phi(\theta_t)-A\|\Delta_t\|_2^2+e_t.
\]
The first-order condition at the comparison point is
$\nabla L_t(z_t)=-\Delta_t^*/\eta_t$. Smoothness then gives
\[
    \|g_t\|_2^2\leq B^2\|\Delta_t^*\|_2^2
       \leq2B^2\|\Delta_t\|_2^2+4B^2e_t/M.
\]
Consequently,
\begin{align*}
    \Phi(\theta_{t+1})
       &\leq\Phi(\theta_t)-\frac{A}{2B^2}\|g_t\|_2^2
                                      +(1+2A/M)e_t\\
       &\leq\Phi(\theta_t)-\frac{\eta_t}{18}\|g_t\|_2^2+3e_t,
\end{align*}
using $A/(2B^2)\geq\eta_t/18$ and $2A/M\leq2$.
Substituting \eqref{eq:empirical-oracle-error} gives
\begin{equation}
\begin{aligned}
    \Phi(\theta_{t+1})\leq{}&\Phi(\theta_t)
       -\frac{\eta_t}{18}\|\nabla\Phi(\theta_t)\|_2^2\\
       &+6\varepsilon_{n,t}+6\delta_m^0+3\varepsilon_{\mathrm{opt},t}.
\end{aligned}
    \label{eq:empirical-descent}
\end{equation}

\subsection{Closing the initial-margin induction}
\label{app:localization-proof}

The preceding one-step bounds apply whenever $\theta_t\in\mathcal B_0$;
they do not require $\theta_{t+1}$ to be in the ball. We now prove that
exit cannot occur, which avoids assuming the desired future margins.
Let $\tau_{\rm exit}=\inf\{t:\theta_t\notin\mathcal B_0\}$.
The conditional uniform-deviation argument applies on each pre-exit
history. A union bound over these histories gives an event of probability
at least $1-\delta$ on which all the required pre-exit bounds hold.
Work on this event.

For $t<\tau_{\rm exit}$, local PL and
\eqref{eq:empirical-descent}, with $\eta_t=18/[\mu(t+s)]$, give
\begin{equation}
    E_{t+1}\leq\left(1-\frac{2}{t+s}\right)E_t+3e_t.
    \label{eq:decreasing-step-recursion}
\end{equation}
If $e_t\leq\zeta/(t+s)^2$, elementary induction gives
\begin{equation}
    E_t\leq\frac{C_0}{t+s-1},\qquad C_0=(s-1)E_0+3\zeta,
    \label{eq:local-potential-envelope}
\end{equation}
up to and including a potential first exit: indeed, for $k=t+s$,
\[
    \left(1-\frac2k\right)\frac{C_0}{k-1}
       +\frac{3\zeta}{k^2}\leq\frac{C_0}{k},
\]
because $C_0\geq3\zeta$.

We also need an \emph{upper} bound on movement, not merely descent.
Smoothness of $L_t$ in \eqref{eq:descent-upper-model} gives, for every
$h\in\mathbb R^d$,
\[
    \Phi(\theta_t+h)\leq\Phi(\theta_t)+\langle g_t,h\rangle
                                     +\frac H2\|h\|^2.
\]
Taking $h=-g_t/H$ and using the global lower bound $\Phi_*$ yields
$\|g_t\|\leq\sqrt{2HE_t}$. This argument uses a touching upper model;
it does not assume that $\Phi$ is globally smooth across modal ties.
The first-order condition for the comparison minimizer $z_t$ implies
\[
    \|z_t-\theta_t\|
       \leq\eta_t\bigl(\|g_t\|+L\|z_t-\theta_t\|\bigr)
       \leq\eta_t\|g_t\|+\tfrac12\|z_t-\theta_t\|.
\]
Together with $\|\theta_{t+1}-z_t\|\leq2\sqrt{\eta_t e_t}$, this gives
\begin{equation}
    \|\theta_{t+1}-\theta_t\|
       \leq2\eta_t\sqrt{2HE_t}+2\sqrt{\eta_t e_t}.
    \label{eq:mode-preserving-step}
\end{equation}
Substituting the error and potential envelopes and summing any prefix,
\begin{align*}
    \sum_{t<j}\|\theta_{t+1}-\theta_t\|
       &\leq\frac{36\sqrt{2HC_0}}{\mu}
          \sum_{t\geq0}\frac1{(t+s)\sqrt{t+s-1}}\\
       &\quad+2\sqrt{\frac{18\zeta}{\mu}}
          \sum_{t\geq0}(t+s)^{-3/2}\\
       &\leq\frac{72\sqrt{2HC_0}}{\mu\sqrt{s-2}}
            +4\sqrt{\frac{18\zeta}{\mu(s-1)}}\leq r_0.
\end{align*}
The integral bounds use $s\geq3$. If a first exit $j\leq T$ existed,
all terms in this prefix would obey the bounds, contradicting
$\|\theta_j-\theta_0\|>r_0$. Hence every empirical iterate stays in
$\mathcal B_0$. Equation~\eqref{eq:initial-gap-preservation} now proves
mode preservation and gap at least $\gamma_0/2$ for all rounds. The
statistical and descent bounds therefore hold throughout the run.

\subsection{Unrolling the recursion and proving the risk bound}

For $\Lambda_T=(T+s-2)(T+s-1)$, the products in
\eqref{eq:decreasing-step-recursion} telescope:
\[
    \prod_{j=t+1}^{T-1}\left(1-\frac2{j+s}\right)
       =\frac{(t+s-1)(t+s)}{\Lambda_T}.
\]
Consequently the fully explicit finite-sample bound is
\begin{equation}
\begin{aligned}
    \Phi(\theta_T)\leq{}&\Phi_*
       +\frac{(s-2)(s-1)E_0}{\Lambda_T}\\
       &+\frac3{\Lambda_T}\sum_{t=0}^{T-1}(t+s-1)(t+s)
          [2\varepsilon_{n,t}+2\delta_m^0+\varepsilon_{\mathrm{opt},t}].
\end{aligned}
    \label{eq:general-nt-bound}
\end{equation}
This is bounded by $B_T$ in \eqref{eq:exact-statistical-bound} under
the assumed complexity envelope. Equation~\eqref{eq:local-potential-envelope}
gives $B_T\leq\Phi_*+C_0/(T+s-1)$ as well, by applying the same
scalar recursion to its deterministic right-hand side. No population
optimization step or output averaging is used.
We next establish the induced adversarial set and its risk. The Gibbs
identity shows that, for any $D$ with
$\KL(D\|\pi_0)\leq\kappa_\beta(\theta)$,
\[
    \E_Dc_\theta
       \leq\Phi(\theta)+\beta\KL(D\|\pi_0)
       \leq\Phi(\theta)+\beta\kappa_\beta(\theta)
       =\E_{q_\beta^\theta}c_\theta.
\]
The last equality follows directly from
$\log(q_\beta^\theta/\pi_0)=c_\theta/\beta-\log Z_\theta$.
Since $q_\beta^\theta$ itself is feasible, it attains the worst-case value.
This proves \eqref{eq:natural-radius} and
\eqref{eq:natural-adversarial-risk}.

For $r\in[0,1]$, concavity and the values at zero and one give
\[
    1-e^{-r/\beta}\geq r(1-e^{-1/\beta}).
\]
Furthermore,
\[
    \E_{q_\beta^\theta}e^{-c_\theta/\beta}
      =\frac1{Z_\theta}=e^{-\Phi(\theta)/\beta}.
\]
Taking expectations in the preceding scalar inequality therefore proves
\begin{equation}
    \operatorname{Unc}_{\kappa_\beta(\theta)}(\theta)
       \leq\frac{1-e^{-\Phi(\theta)/\beta}}{1-e^{-1/\beta}}.
    \label{eq:tilted-to-induced-risk}
\end{equation}
The right-hand side increases with $\Phi(\theta)$. Since
$0\leq\Phi(\theta_T)\leq\min\{B_T,1\}$, substitution gives
\eqref{eq:induced-finite-sample-bound} on the same probability-$1-\delta$
event. This completes the proof of Theorem~\ref{thm:optimal-challenger}.
\hfill$\square$

\subsection{Scope without PL}

Before a possible exit, the one-step argument without PL gives only
\begin{equation}
    \sum_{t<T}\eta_t\|\nabla\Phi(\theta_t)\|^2
       \leq18[\Phi(\theta_0)-\Phi_*]+54\sum_{t<T}e_t,
    \label{eq:inexact-stationarity}
\end{equation}
if all these iterates are separately known to remain in the neighborhood.
It does not establish last-iterate convergence of $\Phi$ or preservation
from the initial margin alone. Theorem~\ref{thm:optimal-challenger}
explicitly retains local PL and the localization budget for both conclusions.

\section{Proof of Corollary~\ref{cor:accuracy}}
\label{app:accuracy}

\paragraph{Modal-answer preservation.}
Theorem~\ref{thm:optimal-challenger}, proved using only the initial gap
and the explicit localization condition, already gives
$a_T(x)=a_0(x)$, $\pi_0$-almost surely. No new small-step condition or
per-round margin assumption is needed for this corollary. It also does
not require every finite-rollout pseudo-label to equal the population mode.

\paragraph{Transferring uncertainty to truth.}
Define the initial error set $A_0=\{x:a_0(x)\ne a^*(x)\}$ and the final
sampling error under a question law $D$ by
\[
    \operatorname{Err}_D(\theta_T)
       :=\Pr_{x\sim D,\,y\sim\pi_{\theta_T}(\cdot\mid x)}
                                       [g(y)\ne a^*(x)]
        =\E_D[1-P_{\theta_T}(a^*(x)\mid x)].
\]
Mode preservation yields the pointwise inequality
\begin{equation}
    1-P_{\theta_T}(a^*(x)\mid x)
       \leq1-p_{\theta_T}(x)+\ind\{x\in A_0\}.
    \label{eq:accuracy-decomposition}
\end{equation}
For every $D$ with $\KL(D\|\pi_0)\leq\kappa_T$, Pinsker's inequality
(with natural logarithms) gives
\[
    D(A_0)\leq\pi_0(A_0)+\operatorname{TV}(D,\pi_0)
            \leq\varepsilon_0+\sqrt{\kappa_T/2}.
\]
Taking expectations in \eqref{eq:accuracy-decomposition} and applying
Theorem~\ref{thm:optimal-challenger} on its simultaneous event proves
\begin{equation}
    \sup_{D:\,\KL(D\|\pi_0)\leq\kappa_T}
       \operatorname{Err}_D(\theta_T)
       \leq\min\{1,\varepsilon_0+b_T+\sqrt{\kappa_T/2}\}.
    \label{eq:robust-error-bound}
\end{equation}
This intermediate bound transfers uncertainty to truth. If $\varepsilon_0=0$,
finite KL implies $D\ll\pi_0$, hence $D(A_0)=0$ for every feasible $D$.
The sharper error bound is then $b_T$.

\paragraph{An explicit radius bound.}
By \eqref{eq:natural-adversarial-risk}, $\Phi_\beta\geq0$, and the theorem,
\begin{equation}
    \beta\kappa_T
       =\E_{q_\beta^{\theta_T}}[1-p_{\theta_T}(x)]-\Phi_\beta(\theta_T)
       \leq b_T.
    \label{eq:accuracy-radius-bound}
\end{equation}
Consequently, a bound involving only $B_T$, $\beta$, and $\varepsilon_0$
is
\begin{equation}
    \inf_{D:\,\KL(D\|\pi_0)\leq\kappa_T}
       \E_D P_{\theta_T}(a^*(x)\mid x)
       \geq\pos{1-\varepsilon_0-b_T-\sqrt{b_T/(2\beta)}}.
    \label{eq:explicit-robust-accuracy}
\end{equation}
\paragraph{Separating the irreducible term from the vanishing error.}
For $v\geq0$, define
\[
    g_\beta(v):=\frac{1-e^{-\min\{v,1\}/\beta}}{1-e^{-1/\beta}},
    \qquad C_\beta:=\frac{1}{\beta(1-e^{-1/\beta})}.
\]
Then $b_T=g_\beta(B_T)$ and $b_*=g_\beta(\Phi_*)$, since
$0\leq\Phi_*\leq1$. For $0<v<1$,
$g_\beta'(v)=e^{-v/\beta}/[\beta(1-e^{-1/\beta})]\leq C_\beta$,
and $g_\beta$ is constant above one. It is therefore nondecreasing and
$C_\beta$-Lipschitz on $[0,\infty)$. With
$\Delta_T=B_T-\Phi_*\geq0$, this gives
\begin{equation}
    b_T\leq b_*+C_\beta\Delta_T.
    \label{eq:accuracy-bstar-comparison}
\end{equation}
Combining \eqref{eq:robust-error-bound},
\eqref{eq:accuracy-radius-bound}, and
$\sqrt{a+b}\leq\sqrt a+\sqrt b$ for $a,b\geq0$ yields
\begin{equation}
\begin{aligned}
    \sup_{D:\,\KL(D\|\pi_0)\leq\kappa_T}
       \operatorname{Err}_D(\theta_T)
    &\leq\min\!\left\{1,\;
       \varepsilon_0+b_*+\sqrt{\frac{b_*}{2\beta}}
       +C_\beta\Delta_T+
          \sqrt{\frac{C_\beta\Delta_T}{2\beta}}\right\}.
\end{aligned}
    \label{eq:accuracy-explicit-remainder}
\end{equation}
This proves \eqref{eq:robust-accuracy-bound}. Its remainder is explicit
in the task counts, labeling rollouts, and optimization errors through
$\Delta_T=B_T-\Phi_*$ in \eqref{eq:exact-statistical-bound}.
Under the theorem's budgets, $\Delta_T\leq C_0/(T+s-1)$, so the remainder
is at most
\[
    \frac{C_\beta C_0}{T+s-1}
       +\sqrt{\frac{C_\beta C_0}{2\beta(T+s-1)}}
       =O_\beta(T^{-1/2}).
\]
For $\Phi_*=0$, $b_*=0$ and only $\varepsilon_0$ remains in the limiting upper bound.
For $\varepsilon_0=0$, the earlier bound $\operatorname{Err}_D(\theta_T)
\leq b_T$ removes the distribution-shift term entirely; then
\eqref{eq:accuracy-bstar-comparison} gives $b_*+C_\beta\Delta_T$,
with an $O_\beta(1/T)$ remainder under the same budgets.
These are bounds on the induced KL set for each $T$, not on a fixed-radius
set, and $b_*+\sqrt{b_*/(2\beta)}$ is a consequence of the optimal
uncertainty potential, not an optimal true-error value.

Under the base question law itself, no distribution-shift term is needed:
$\operatorname{Err}_{\pi_0}(\theta_T)\leq\min\{1,\varepsilon_0+b_T\}$.
Here $\varepsilon_0$ bounds the initial \emph{population modal-answer}
error, not its single-response error. The corollary concerns sampling
accuracy, not a change in modal predictions, and it does not certify
the practical triangular-reward GRPO implementation.
\hfill$\square$

\section{Extension to Approximate Question Sampling}
\label{app:sampler-error}

Theorem~\ref{thm:optimal-challenger} assumes the exact optimal question law
while retaining finite-$m$ labels and finite-$n_t$ empirical training. However in practice we can only estimate $p_\theta$ with finite roll-outs. Here we examine the bound under this realization.
For a sampler law $\widetilde q_t$ with
$\operatorname{TV}(\widetilde q_t,q_t)\leq d_t$, retain conditional iid
questions, fresh independent labels, and $\widetilde q_t\ll\pi_0$ so that
the initial almost-sure margin applies to its support. Bounded losses
add at most $d_t$ to \eqref{eq:empirical-loss-comparison}; replace
$e_t$ by $e_t+2d_t$ everywhere, including the error schedule and
localization condition. Under these augmented conditions,
\begin{equation}
    B_T^{\mathrm{approx}}:=B_T
       +\frac6{\Lambda_T}\sum_{t=0}^{T-1}(t+s-1)(t+s)d_t.
    \label{eq:approximate-sampler-bound}
\end{equation}
The complexity is evaluated under the actual question--label law.
Both $q_T$ and the induced radius still refer to the ideal challenger of
the final model. Sampler error consumes the same localization budget;
it cannot simply be added to the final bound while ignoring its effect
on preservation. Constant nonzero sampler error is not covered by the
vanishing-error asymptotic rate.

\paragraph{Expected finite-rollout reward.}
\label{app:finite-m}
For the theoretical uncertainty reward $r_c=1-p_\theta$ in
\eqref{eq:concentration-rewards}, define
$\rho_{\theta,m}(x)=\E[1-\max_a\widehat P_m(a\mid x)]$, where
$\widehat P_m$ uses $m$ independent valid answers. Convexity of the maximum
and the multinomial frequency error give
\begin{align*}
    0\leq r_c(x,\theta)-\rho_{\theta,m}(x)
    &\leq\E\|\widehat P_m-P_\theta\|_\infty\\
    &\leq\sqrt{\frac{1-\sum_aP_\theta(a\mid x)^2}{m}}
      \leq\frac1{\sqrt m}.
\end{align*}
If a sampler exactly targets
$q_t^{(m)}\propto\pi_0e^{\rho_{\theta_t,m}/\beta}$, the oscillation of its
log-density ratio with $q_t$ is at most $h=1/(\beta\sqrt m)$, implying
\begin{equation}
    d_t\leq\tanh\!\left(\frac1{4\beta\sqrt m}\right).
    \label{eq:finite-m-target-error}
\end{equation}
To see this, a density ratio $w\in[a,b]$ of mean one with $b/a\leq e^h$
satisfies
$\operatorname{TV}\leq(b-1)(1-a)/(b-a)$ by convexity of $|w-1|$.
Maximizing over $a\leq1\leq b$ gives $\tanh(h/4)$.

This particular bound assumes the exponential of the expected finite-rollout
reward and independent labeling responses. Exponentiating one noisy score,
reusing selection responses as labels, filtering the question law, or using
a diversity-coupled batch needs further analysis. These conditions are not
an error certificate for the practical walk in Appendix~\ref{app:practical-details}.

\section{Mathematical-Reasoning Implementation}
\label{app:practical-details}

This appendix instantiates the general algorithm of Section~\ref{sec:framework}
for the mathematical-reasoning experiments. The concrete reward, LLM
mutation instructions, finite-step acceptance rule, and solver training
choices are collected here. This practical configuration differs from
the ideal-sampling $1-p$ and L2 instance analyzed in Section~\ref{sec:theory}.

\begin{table}[!htbp]
\centering
\small
\caption{Configuration of the local-generator DEO runs. The agreement
rollouts and GRPO training rollouts serve different purposes.}
\label{tab:setup}
\resizebox{\linewidth}{!}{\begin{tabular}{@{}ll@{}}
\toprule
Setting & Value \\
\midrule
Base models & Qwen3-4B/8B-Base; OctoThinker-3B-Hybrid-Base \\
Candidate pool per round & $N=2000$, freshly generated \\
Mutation sweeps & $L_{\mathrm w}=5$ \\
MH temperature & Fixed $\beta=0.1$ \\
Agreement rollouts per evaluated question & $m=9$ \\
Agreement sampling & Temperature $1.0$, top-$p=1.0$, top-$k=40$ \\
Solver response limit & 4096 tokens \\
BLEU distance threshold / repetition weight & $0.5$ / $\lambda_{\mathrm{rep}}=10$ \\
Training agreement filter & $\widehat p\in[0.3,0.8]$ \\
Solver training & GRPO; 5 training responses per prompt \\
Solver KL reference / coefficient & Original base model / $0.01$ \\
Additional sampling interventions & None \\
Reported checkpoints & Iterations 1, 2 and 3 \\
\bottomrule
\end{tabular}}
\end{table}

\subsection{Concrete challenger reward}

For each question $x$, draw $m$ responses from the frozen current solver.
Let $\widehat p(x)$ be the largest frequency of an extracted answer,
divided by $m$. We use
\begin{equation}
    \widehat r_c(x,\theta)=1-2\left|\widehat p(x)-\tfrac12\right|.
    \label{eq:uncertainty-reward}
\end{equation}
For a pool $X=(x_1,\ldots,x_N)$, the score used for selection is
\begin{equation}
    \widehat r_c(X,\theta)=\sum_{i=1}^N
       [\widehat r_c(x_i,\theta)
                     -\lambda_{\mathrm{rep}}r_{\mathrm{rep}}(x_i;X)]_+,
    \qquad r_{\mathrm{rep}}(x_i;X)=\frac{|\mathcal N_i(X)|}{N},
    \label{eq:empirical-batch-utility}
\end{equation}
where $\mathcal N_i(X)$ contains the BLEU-based neighbors of $x_i$,
including itself. We use $m=9$ and $\lambda_{\mathrm{rep}}=10$.
The same notation $r_c$ applies to individual questions and complete pools;
hats indicate cached finite-rollout estimates. The answer-extraction and neighbor-count definitions follow below.

\subsection{Answer statistics and repetition}

The frozen pretrained model $\pi_{\mathrm{base}}$ induces the fixed seed
question law $\pi_0$ through its generation prompt, predefined topic
sampling, and output parsing. During each walk, the current solver is frozen.
For responses $\xi=(y_1,\ldots,y_m)$, let $g(y)=\bot$ denote an invalid
answer extraction and define, for $a\ne\bot$,
\begin{equation}
    c_a=\sum_{j=1}^m\ind\{g(y_j)=a\},\qquad
    \widehat p=\frac{\max_{a\ne\bot}c_a}{m},\qquad
    \widehat a\in\arg\max_{a\ne\bot}c_a.
    \label{eq:modal-answer}
\end{equation}
If no answer is extracted, set $\widehat p=0$ and $\widehat a=\bot$.
Invalid responses remain in the denominator $m$. Answers are grouped by
extracted string, without full mathematical-equivalence clustering.
Thus $\widehat p$ is empirical answer agreement, not verified correctness.
The stored pseudo-label comes from the current solver, not from the
mutator's proposed solution.

The symmetric neighbor matrix thresholds smoothed sentence-BLEU distances
at $0.5$. Its row counts include self-neighbors and give
\begin{equation}
    r_{\mathrm{rep}}(x_i;X)=|\mathcal N_i(X)|/N.
    \label{eq:repetition-penalty}
\end{equation}
These are local neighbor counts, not transitive cluster sizes. Clipping
in \eqref{eq:empirical-batch-utility} occurs before the sum. A mutation
changes the selected question's neighbor row and column and the penalties
of all affected neighbors. Unchanged questions retain their uncertainty
estimates. The reported settings are $N=2000$, $m=9$, $L_{\mathrm w}=5$,
$\beta=0.1$, and $\lambda_{\mathrm{rep}}=10$.

\subsection{LLM mutation and the implemented walk}
\label{app:llm-walk}

Each round starts with $N$ seed questions from the frozen base LLM.
For each current question $x_k$, we prompt this LLM to produce a structurally
different question $x_k'$ by choosing one of five operations:
generalize the setting, compose interacting conditions, invert the task,
change the requested quantity, or exchange a mathematical concept for its
dual. The prompt discourages mere number substitution. Thus mutation is
\emph{conditional text generation by an LLM}, rather than a learned
challenger update. Denote this prompted proposal kernel by
$K_{\mathrm{LLM}}(x_k'\mid x_k)$.

The current solver draws $m$ fresh responses to $x_k'$; these provide both
its uncertainty score and its proposed pseudo-label. Let $X'$ replace
$x_k$ by $x_k'$. We accept with probability
\begin{equation}
    \widehat\alpha(X,X')=\min\left\{1,
      \exp\left(\frac{\widehat r_c(X',\theta_t)
                          -\widehat r_c(X,\theta_t)}{\beta}\right)\right\}.
    \label{eq:implemented-acceptance}
\end{equation}
The score difference includes all affected repetition penalties.
Acceptance replaces the question and its cached answer statistics;
rejection retains the old state. Unchanged questions reuse their responses.
We perform $L_{\mathrm w}=5$ randomly ordered sweeps at fixed $\beta=0.1$.

This rule is motivated by the KL-optimal batch tilt
$Q_t(X)\propto\Pi_0(X)\exp(r_c(X,\theta_t)/\beta)$, where
$\Pi_0=\pi_0^{\otimes N}$ and $r_c(X,\theta_t)=\E\widehat r_c(X,\theta_t)$.
The implemented walk is approximate: it uses noisy scores, omits the
base-density/proposal ratio, and runs finitely many steps.
Appendix~\ref{app:practical-target} gives the exact target and acceptance
rule. No challenger training, CD correction, or temperature adaptation is used.

\begin{figure}[t]
\centering
\begin{DEOalgorithm}
  {LLM-based mutation and Metropolis selection}
  {algorithm:fixed-beta-walk}
\textbf{Input:} initial pool $X^0$, frozen current solver $\pi_\theta$,
frozen LLM mutation kernel $K_{\mathrm{LLM}}$, temperature $\beta$,
sweeps $L_{\mathrm w}$, rollout count $m$.\par
\textbf{Output:} final pool $X$ and cached $(\widehat p_i,\widehat a_i)_{i=1}^N$.
\begin{algorithmic}[1]
\State Set $X\leftarrow X^0$; draw $m$ responses per question and cache
  $(\widehat p_i,\widehat a_i,\widehat r_c(x_i,\theta))$.
\For{$\ell=1,\ldots,L_{\mathrm w}$}
    \State Draw a random permutation $\sigma$ of $\{1,\ldots,N\}$.
    \For{$k=\sigma(1),\ldots,\sigma(N)$}
        \State Prompt the frozen LLM with $x_k$ to sample
          $x_k'\sim K_{\mathrm{LLM}}(\cdot\mid x_k)$.
        \If{$x_k'$ is parseable}
            \State Draw $m$ fresh solver responses to $x_k'$;
              compute its answer statistics.
            \State Form $X'\leftarrow X[k\leftarrow x_k']$ and compute
              $\Delta\leftarrow\widehat r_c(X',\theta)-\widehat r_c(X,\theta)$.
            \State Accept $X'$ and its statistics with probability
              $\min\{1,e^{\Delta/\beta}\}$; otherwise retain the cached state.
        \EndIf
    \EndFor
\EndFor
\State \Return $X$ and its cached answer statistics.
\end{algorithmic}
\end{DEOalgorithm}
\end{figure}

\subsection{Batch target and approximation scope}
\label{app:practical-target}

Write $\widehat r_c(X,\theta;\xi)$ when the rollout randomness needs to be
explicit. The ideal batch reward is its finite-rollout expectation,
\begin{equation}
    r_c(X,\theta)=\E_\xi\widehat r_c(X,\theta;\xi).
    \label{eq:batch-utility}
\end{equation}
It is not the triangular function applied to population probabilities.
With $\Pi_0=\pi_0^{\otimes N}$, the batch challenger objective and optimizer are
\begin{align}
    \mathcal J_t(Q)&=\E_Qr_c(X,\theta_t)-\beta\KL(Q\|\Pi_0),
    \label{eq:batch-challenger-obj}\\
    Q_t^*(X)&=\frac{\Pi_0(X)e^{r_c(X,\theta_t)/\beta}}{Z_t},\qquad
    Z_t=\E_{\Pi_0}e^{r_c(X,\theta_t)/\beta}.
    \label{eq:batch-target}
\end{align}
Indeed,
\begin{equation}
    \mathcal J_t(Q)=\beta\log Z_t-\beta\KL(Q\|Q_t^*).
    \label{eq:gibbs-identity}
\end{equation}
This optimizes over unrestricted batch laws. The repetition term couples
questions, so the optimum generally does not factorize.

Let $\mathcal K$ be the batch proposal kernel induced by coordinate selection and
$K_{\mathrm{LLM}}$. Exact Metropolis--Hastings acceptance would require
\begin{equation}
    \alpha^*(X,X')=\min\left\{1,
       \frac{\Pi_0(X')\mathcal K(X\mid X')}{\Pi_0(X)\mathcal K(X'\mid X)}
       e^{[r_c(X',\theta_t)-r_c(X,\theta_t)]/\beta}\right\}.
    \label{eq:exact-acceptance}
\end{equation}
The implementation omits the base-density/proposal ratio and uses cached
noisy scores. The omitted factor cancels if $\mathcal K$ is reversible with respect
to $\Pi_0$; this is not established for the prompted LLM mutator.
Symmetry of $\mathcal K$ alone is insufficient. Also,
$\E e^{\widehat r_c/\beta}\ne e^{\E\widehat r_c/\beta}$ in general,
and a finite walk need not mix. These distinctions prevent interpreting
the implemented walk as exact sampling from \eqref{eq:batch-target}.
No CD penalty, U-statistic correction, adaptive temperature, memory,
or mutation bandit is used in the reported configuration.

The frozen LLM uses the standard V1 prompt: it chooses one of generalize,
compose, invert, change-objective, and dualize, conditioned on the current
question. Proposals are generated in batches for throughput, with their
score changes processed against the current pool. The prompt requests a
number, algebraic expression, or finite-set answer, disallows proof and
yes/no questions, and rejects mere number swapping as a mutation strategy.
The exact generation, mutation, solver, and evaluation prompts are given
in Appendix~\ref{app:prompt-templates}.

\subsection{Format filtering and GRPO}
\label{app:practical-filter}

The final dataset contains only pairs with $\widehat p\in[0.3,0.8]$
and a nonempty pseudo-label that pass the implemented format checks.
The code applies parsing/regex checks and an LLM \texttt{VALID}/\texttt{INVALID}
judge for template text, prompt leakage, corrupted output, non-math or
unsupported question types, and missing answers. The judge is explicitly
instructed not to solve the problem; it is not a correctness verifier.
For $m=9$, the band keeps modal counts $3$ through $7$. The budget $N=2000$
is measured before filtering, so the retained training size varies by round.

The GRPO implementation maximizes
\begin{equation}
    R_{\mathrm{GRPO}}(x,y;\widehat a)
       =0.9\,\ind\{\operatorname{grade}(g(y),\widehat a)=1\}
         +0.1\,r_{\mathrm{format}}(y),
    \label{eq:solver-reward}
\end{equation}
where the grader checks answer equivalence and $r_{\mathrm{format}}$
checks the inherited reasoning/final-answer format. Equivalently, it
minimizes the response-averaged loss
\begin{equation}
    0.9\,r_s(x,\theta;\widehat a)
       +0.1\,\E_{y\sim\pi_\theta(\cdot\mid x)}[1-r_{\mathrm{format}}(y)],
    \label{eq:practical-solver-loss}
\end{equation}
with $r_s(x,\theta;\widehat a)$ interpreted using the same answer-equivalence
grader. Here \[r_s(x,\theta;\widehat a)=1-\Pr_{y\sim\pi_\theta(\cdot\mid x)}[\operatorname{grade}(g(y),\widehat a)=1].\]
KL regularization has coefficient $0.01$ and uses the original base model
as reference; this differs from the previous-iterate L2 instance of the
theorem. Training draws fresh responses against the stored pseudo-labels
and returns the checkpoint specified in the experimental protocol.

\section{Experimental Protocol and Supplementary Results}
\label{app:experimental-protocol}
\subsection{Experimental Protocol}
\paragraph{Models and question generation.}
We use Qwen3-4B-Base, Qwen3-8B-Base and
OctoThinker-3B-Hybrid-Base. For the local-generator runs, the generator
and initial solver use the same checkpoint. DEO starts
each round with $N=2000$ parsed candidate questions and applies five
mutation sweeps at a fixed $\beta=0.1$. Initial questions are sampled
with temperature $1.0$ and top-$p=0.95$; the standard V1 mutator uses
temperature $1.1$. Both generation calls have a 1536-token output limit.
The generator is frozen at the original pretrained checkpoint. The challenger reward and BLEU repetition penalty are defined in
Appendix~\ref{app:practical-details}, with no additional intervention. Table~\ref{tab:setup} gives the
shared settings. Section~\ref{sec:claude-generator} additionally replaces
the 4B generator and mutator with Claude while retaining the Qwen solver.

\paragraph{Solver training and comparisons.}
The native training jobs use the inherited verl GRPO configuration with
64 prompts per rollout batch, five responses per prompt, learning rate
$10^{-6}$ and a configured maximum of 20 training steps per round. The
runner exports and evaluates the \texttt{global\_step\_15} checkpoint,
and uses that checkpoint to initialize the next round. Thus the configured
20-step job budget should not be confused with the checkpoint index.
Experiments run on an eight-H100 AzureML node, with separate resources for
generation, solver scoring, training and evaluation. The no-walk ablation
also generates 2000 candidates per round and uses the same solver labeling,
filtering and training recipe, but sets $L_{\mathrm w}=0$. This comparison holds the
candidate-pool and solver-update configurations fixed; it does not equalize
the additional generation and scoring cost of the walk.

We also include the archived R-Zero~\citep{huang2025r} runs for all three
base models. R-Zero trains a separate challenger with RL and uses
solver-produced modal answers to supervise solver training. Its solver
uses the same nominal GRPO update budget and exported checkpoint index;
we compare iterations one through three under the same evaluation protocol.
The 2000-candidate restriction applies to DEO and no walk: the Qwen R-Zero runs retain
their native, larger question-generation pools. For example, its first three
4B training sets contain 3939, 4312 and 4376 post-filter questions.
R-Zero is therefore a method-level baseline, whereas no walk is the
matched-pool ablation. This comparison does not establish equal total
compute or attribute every difference solely to the walk.

\paragraph{Evaluation.}
For mathematical benchmarks, we report the first three iterations from
the archived runs using the same Claude Haiku boxed-answer grading protocol.
AVG7 is the unweighted mean over MATH-500, GSM8K, AMC, Minerva,
OlympiadBench, AIME 2024, and AIME 2025; HARD5 averages the last five.
All entries are percentages. Main-table means and the differences discussed
in the text are calculated from the reported per-benchmark percentages
and rounded after calculation. Omni-MATH is evaluated separately in
Appendix~\ref{app:omni-math} and is excluded from AVG7 and HARD5.
The evaluation loader repeats AMC and AIME questions 32 times; these
repetitions are not additional independent problems. We therefore report
the observed scores without treating repeated rows as independent samples
for statistical significance. These mathematical comparisons use a
common grading protocol without checkpoint selection beyond iteration three.

For general-domain transfer, we evaluate all three base models on MMLU-Pro,
SuperGPQA and BBEH using fixed stratified subsets (target size 3000 per
benchmark; seed 0), shared across checkpoints. Generation is greedy with
at most 4096 new tokens. Multiple-choice answers are graded by option
letter, and BBEH uses normalized exact match; this evaluation does not
use an LLM judge. AVG3 is the unweighted mean of the three accuracies.
This is our project's evaluation protocol, rather than a reproduction
of R-Zero's original general-domain evaluation harness.

\paragraph{Closed-source generation.}
The Claude run uses Claude Haiku 4.5 for both seed generation and mutation,
with the Qwen3-4B solver unchanged. The API uses a generation temperature
capped at 1.0, whereas the local V1 mutator uses 1.1. Both runs retain
the same candidate budget, solver-scoring and labeling roles, filtering,
training, and evaluation conventions. Claude responses do not supply
the solver's target labels.

\paragraph{Candidate pools and retained datasets.}
The 2000-task budget is measured before the final filter. The archived
DEO datasets contain 1246, 1208, and 1228 retained training tasks in the
first three 4B rounds, and 1520, 1508, and 1510 at 8B. These are dataset
sizes, not counts of distinct tasks guaranteed to be consumed by the
fixed-length solver-training job.

\subsection{Additional evaluation on Omni-MATH}
\label{app:omni-math}
We additionally evaluate the base and iteration-three checkpoints on
Omni-MATH~\citep{gao2024omnimath}, using a fixed 2000-question subset and
the same harness across all models and methods. These scores are reported
separately from the seven-benchmark comparison in the main text and do
not enter AVG7 or HARD5.

\begin{table}[!htbp]
\centering
\small
\setlength{\tabcolsep}{7pt}
\caption{Omni-MATH accuracy (\%) on the fixed 2000-question subset.
All trained methods use iteration
three.   Bold marks the best trained-method score in each row.}
\label{tab:omni-math}
\begin{tabular}{@{}lrrrrr@{}}
\toprule
Model & Base & No walk & R-Zero & DEO \\
\midrule
OctoThinker-3B & 9.50  & 12.80 & \textbf{13.10} & 12.80  \\
Qwen3-4B & 22.00 & \textbf{25.50} & 24.75 & 24.80  \\
Qwen3-8B & 23.55 & 26.55 & 26.65 & \textbf{27.00} \\
\bottomrule
\end{tabular}
\end{table}

Table~\ref{tab:omni-math} shows similar accuracy across the trained methods,
with model-dependent rankings: no walk leads at 4B, DEO leads at 8B,
and R-Zero leads on OctoThinker-3B.  

\subsection{First-three-round trajectories}
Table~\ref{tab:iteration-results} gives the archived trajectories.
The iteration-three rows agree with the primary comparison; intermediate
rankings can differ.

\begin{table}[!htbp]
\centering
\small
\setlength{\tabcolsep}{7pt}
\caption{DEO, R-Zero and the no-walk ablation through iteration three.
DEO and no walk use 2000 initial candidates per round; R-Zero uses its
native larger question pool. DEO uses fixed $\beta=0.1$, five sweeps
and plain noisy acceptance, without CD or auxiliary interventions.
Bold marks the best score among the three methods at each scale and iteration.}
\label{tab:iteration-results}
\begin{tabular}{@{}llrrrrrr@{}}
\toprule
& & \multicolumn{3}{c}{AVG7 ($\uparrow$)}
    & \multicolumn{3}{c}{HARD5 ($\uparrow$)} \\
\cmidrule(lr){3-5}\cmidrule(l){6-8}
Model & Method & Iter 1 & Iter 2 & Iter 3 & Iter 1 & Iter 2 & Iter 3 \\
\midrule
4B & No walk & 45.49 & 47.73 & 45.68
   & 30.02 & 32.84 & 30.49 \\
   & R-Zero & \textbf{46.46} & \textbf{48.13} & 45.93
   & \textbf{31.38} & \textbf{33.96} & 31.16 \\
   & DEO & 45.09 & 47.71 & \textbf{47.57}
   & 29.02 & 33.18 & \textbf{32.96} \\
\midrule
8B & No walk & 51.91 & 50.64 & 51.78 & 38.04 & 36.28 & 37.49 \\
   & R-Zero & 51.96 & 51.70 & 52.88 & 37.88 & 37.68 & 38.97 \\
   & DEO & \textbf{52.01} & \textbf{53.39} & \textbf{53.60}
   & \textbf{38.22} & \textbf{40.02} & \textbf{39.96} \\
\bottomrule
\end{tabular}
\end{table}

\clearpage
\section{Prompt Templates}
\label{app:prompt-templates}

We give the prompt templates used in the reported experiments. The panels
separate message roles, following the presentation of R-Zero~\citep{huang2025r}.
Text is reproduced from the implementation, with line wrapping adjusted
for readability. Placeholder meanings are given below each panel; chat messages
are serialized with the corresponding model's chat template.
The seed-generation and V1 mutation templates are shared by the local
and Claude generators. No walk uses the same seed-generation and solver
prompts, without mutation.

\subsection{Initial task generation}
\label{app:prompt-seed}
\begin{DEOprompt}{Seed-generation prompt}
\textbf{System Message:}\par\smallskip
You are an expert competition-math problem setter. FIRST, in your private scratch-pad, think step-by-step to design a brand-new, non-trivial problem. Aim for a medium-to-hard competition level.\par
\smallskip
CRITICAL RULES:\par
1. The final answer MUST be a SPECIFIC NUMBER, ALGEBRAIC EXPRESSION, or FINITE SET.\par
2. DO NOT generate "Prove that", "Show that", "Justify", or "Explain why" questions.\par
3. DO NOT generate questions that ask for True/False or Yes/No answers.\par
4. Ensure the problem has exactly one unambiguous final answer.\par
5. LIMIT YOUR SCRATCH-PAD THINKING TO UNDER 50 WORDS! Do not write out the full proof.\par
\smallskip
THEN, output the problem and answer exactly in this format:\par
\texttt{\textless{}question\textgreater{}}\par
[Write the full problem statement here on one or more lines]\par
\texttt{\textless{}/question\textgreater{}}\par
\texttt{\textbackslash{}boxed\{final\_answer\}}\par
\medskip
\textbf{User Message:}\par\smallskip
Generate one new, challenging reasoning question now. YOU MUST STRICTLY FOCUS ON: **\texttt{\{topic\}}**.\par
\end{DEOprompt}

\noindent\textit{Placeholder.} \texttt{\{topic\}} is sampled uniformly from:
Algebra: Polynomial roots and coefficients;
Geometry: Triangle centers and circles;
Geometry: 3D spatial geometry and volume;
Geometry: Coordinate geometry and loci;
Number Theory: Diophantine equations;
Number Theory: Modular arithmetic and congruences;
Combinatorics: Probability and expected value;
and Calculus: Limits and derivatives.
The generator's boxed answer is not used as the solver's training label.

\subsection{Solver responses}
\label{app:prompt-solver}
\begin{DEOprompt}{Shared solver prompt}
\textbf{System Message:}\par\smallskip
Please reason step by step, and put your final answer within \texttt{\textbackslash{}boxed\{\}}.\par
\medskip
\textbf{User Message:}\par\smallskip
\texttt{\{problem\}}\par
\end{DEOprompt}

\noindent\textit{Use.} \texttt{\{problem\}} is the question being scored,
used for GRPO training, or evaluated on a mathematical benchmark.
The same system message is used for the $m=9$ scoring responses,
modal-answer pseudo-labeling, fresh training responses, and benchmark
evaluation. Training compares fresh answers with the stored pseudo-label;
the pseudo-label is not inserted into this prompt.

\clearpage
\subsection{LLM-based task mutation}
\label{app:prompt-mutation}
The standard V1 template below is used throughout the reported DEO runs.
It asks the frozen LLM to select one structural mutation.
\begin{DEOprompt}{Task-mutation prompt (V1)}
\textbf{System Message:}\par\smallskip
You are an expert competition-math problem setter. I will provide a seed problem.\par
Your task is to generate a STRUCTURALLY DIFFERENT problem by applying ONE of the following mutation strategies.\par
\smallskip
MUTATION STRATEGIES (pick exactly ONE that is NOT already exemplified by the seed):\par
[A] GENERALIZE --- Lift the structure: replace a specific constant with a parameter, OR raise the dimension (2D \(\rightarrow\) 3D, single-variable \(\rightarrow\) multivariable, single equation \(\rightarrow\) system of equations).\par
[B] COMPOSE   --- Add a NEW non-trivial second condition that interacts with the existing structure (NOT just a range bound like "with x \textgreater{} 0"). The two conditions together must create a real interaction.\par
[C] INVERT    --- Given the original answer or output, ask the reader to recover an input or precondition.\par
[D] CHANGE\_OBJECTIVE --- Change WHAT is being asked: e.g. "find x" \(\rightarrow\) "count integer solutions", "compute" \(\rightarrow\) "find the smallest n such that ...", "find value" \(\rightarrow\) "find sum of all such values".\par
[E] DUALIZE   --- Swap to a dual concept: sum\(\leftrightarrow\)product, max\(\leftrightarrow\)min, area\(\leftrightarrow\)perimeter, gcd\(\leftrightarrow\)lcm, addition\(\leftrightarrow\)multiplication, distance\(\leftrightarrow\)angle.\par
\smallskip
CRITICAL RULES:\par
1. DO NOT just swap numbers. If the only edit is a digit change, you have FAILED --- restart with another strategy.\par
2. DO NOT pick the strategy already exemplified by the seed. (e.g. if seed is already 3D, don't pick GENERALIZE\(\rightarrow\)add dimension.)\par
3. The final answer MUST be a SPECIFIC NUMBER, ALGEBRAIC EXPRESSION, or FINITE SET.\par
4. NO "Prove that", "Show that", "Justify", "True/False", or "Yes/No" questions.\par
5. LIMIT scratch-pad reasoning to UNDER 50 WORDS.\par
\smallskip
Output format (STRICT --- all three tags required):\par
\texttt{\textless{}strategy\textgreater{}}\texttt{\{A\textbar{}B\textbar{}C\textbar{}D\textbar{}E\}}\texttt{\textless{}/strategy\textgreater{}}\par
\texttt{\textless{}question\textgreater{}}\par
[the NEW mutated problem statement]\par
\texttt{\textless{}/question\textgreater{}}\par
\texttt{\textbackslash{}boxed\{final\_answer\}}\par
\medskip
\textbf{User Message:}\par\smallskip
Here is the seed problem:\par
\texttt{\{seed\}}\par
\smallskip
Pick ONE mutation strategy from \texttt{\{A,B,C,D,E\}} and apply it now. Remember: number-swapping is FAILURE.\par
\end{DEOprompt}

\noindent\textit{Placeholder.} \texttt{\{seed\}} is the current accepted
question at the selected pool coordinate. Each proposal receives this
question alone: solver responses, numerical scores, and weakness-memory
notes are not included. Solver scoring and the acceptance decision occur
after the proposal is generated. On rejection, the current question is
retained for the next mutation attempt.

\clearpage
\subsection{Training-data format check}
\label{app:prompt-filter}
After parsing and band filtering, the implementation uses the following
completion prompt to screen the question and its solver-produced pseudo-label.
\begin{DEOprompt}{Format-check prompt}
\textbf{Raw Completion Prompt:}\par\smallskip
You are a strict validity filter for competition-math problems.\par
\smallskip
Mark a problem as INVALID ONLY IF you are CONFIDENT one of these specific issues applies:\par
- Contains placeholder/template text like "[the NEW mutated problem statement]", "final\_answer", or "[Write the full problem statement]".\par
- Contains visible prompt leakage, role tags (Assistant:, User:), HTML/XML tags (\texttt{\textless{}question\textgreater{}}, \texttt{\textless{}strategy\textgreater{}}), or markdown code fences (\textasciigrave{}\textasciigrave{}\textasciigrave{}).\par
- Contains corrupted Unicode garbage, randomly-mixed unrelated languages, or web/chat artifacts.\par
- Is clearly NOT a math problem.\par
- Asks for a proof, justification, true/false, yes/no, or open-ended explanation.\par
- The proposed boxed answer is missing or empty.\par
\smallskip
When in doubt, the problem is VALID. Do NOT solve the problem. Do NOT reject for being hard, unusual, or having minor formatting quirks.\par
\smallskip
Problem:\par
\textless{}\textless{}\textless{}\par
\texttt{\{q\}}\par
\textgreater{}\textgreater{}\textgreater{}\par
\smallskip
Proposed boxed answer: \texttt{\{ans\}}\par
\smallskip
Verdict ("VALID" or "INVALID"):\par
\end{DEOprompt}

\noindent\textit{Placeholders and decoding.} \texttt{\{q\}} is the parsed
question and \texttt{\{ans\}} is the solver's modal-answer pseudo-label.
This is sent to the solver endpoint as a raw completion, with temperature
0 and a six-token output limit. It checks format and question type,
not mathematical correctness.

\subsection{Mathematical-answer grading}
\label{app:prompt-grader}
For the reported seven-benchmark results, the boxed-answer regrader uses
Claude Haiku 4.5 (\texttt{claude-\allowbreak{}haiku-\allowbreak{}4-\allowbreak{}5-\allowbreak{}20251001}), with temperature
0.1 and a four-token output limit. No system message is supplied.
\begin{DEOprompt}{Boxed-answer regrading prompt}
\textbf{User Message:}\par\smallskip
Hi, there is an answer: \texttt{\{model\_answer\}},and the ground truth answer is: \texttt{\{reference\_answer\}},please check whether the answer is correct or not, and return the **only**Yes or No.\par
\end{DEOprompt}

\noindent\textit{Placeholders and use.} \texttt{\{model\_answer\}} is the
last boxed answer in the model response, or the literal string
\texttt{(NO BOXED ANSWER FOUND IN MODEL RESPONSE)} if none is found.
\texttt{\{reference\_answer\}} is the last boxed reference answer, falling
back to the reference text. Only responses scored incorrect by the
initial grader are rechecked; an affirmative verdict promotes their
score. The regrader receives these answer strings, not the question or
the full model reasoning. This evaluation judge is separate from the
training-data format check above.

\clearpage
\subsection{General-domain evaluation}
\label{app:prompt-general}
MMLU-Pro, SuperGPQA and BBEH use the shared solver system message
from Appendix~\ref{app:prompt-solver}. Their user messages are shown below.
\begin{DEOprompt}{General-domain evaluation prompts}
\textbf{User Message: MMLU-Pro / SuperGPQA:}\par\smallskip
\texttt{\{question\}}\par
\smallskip
Options:\par
\texttt{\{options\}}\par
\smallskip
Answer with the letter of the single correct option, and put that letter within \texttt{\textbackslash{}boxed\{\}}.\par
\medskip
\textbf{User Message: BBEH:}\par\smallskip
\texttt{\{question\}}\par
\smallskip
Put your final answer within \texttt{\textbackslash{}boxed\{\}}.\par
\end{DEOprompt}

\noindent\textit{Placeholders and use.} \texttt{\{question\}} is the
benchmark question (the input field for BBEH).
\texttt{\{options\}} expands to all supplied choices, one per line,
in the form \texttt{A. option text}, \texttt{B. option text}, and so on.
Generation is greedy, with a 4096-token output limit. Multiple-choice
outputs are graded by option letter and BBEH by normalized exact match;
neither uses the boxed-answer LLM regrader.
Omni-MATH uses the shared solver prompt with its problem statement
and no additional user-message suffix.

\end{document}

%% file: command.tex
\usepackage{epsf}
\usepackage{graphics}
\usepackage{wrapfig}
\usepackage{psfrag}

\usepackage{color}

\usepackage{mathtools}
\usepackage{amsfonts}
\usepackage{amsthm}
\usepackage{amsmath}
\usepackage{amssymb}



\newcommand{\abs}[1]{\left|#1\right|}





\newtheorem{theorem}{Theorem}

\newtheorem{corollary}{Corollary}





\long\def\comment#1{}





\providecommand{\E}{\ensuremath{{\mathbb{E}}}}



\newcommand{\2}{\ensuremath{{\sf (ii)}}}

\newcommand{\5}{\ensuremath{{\sf (v)}}}

\DeclareMathOperator{\var}{var}
\ifdefined\Var\else \DeclareMathOperator{\Var}{Var}\fi

\ifdefined\abs\else \DeclareMathOperator{\abs}{abs}\fi
\ifdefined\floor\else \DeclarePairedDelimiter\floor{\lfloor}{\rfloor}\fi
\ifdefined\ceil\else \DeclarePairedDelimiter{\ceil}{\lceil}{\rceil}\fi

\ifdefined\sign\else \DeclareMathOperator{\sign}{sign}\fi

%










\ifdefined\var\else \DeclareMathOperator{\var}{\mathsf{var}}\fi








\providecommand{\R}{\mathbb{R}}

\newcommand{\cF}{\mathcal{F}}  



\ifdefined\argmax\else \DeclareMathOperator*{\argmax}{arg\,max}\fi
\ifdefined\argmin\else \DeclareMathOperator*{\argmin}{arg\,min}\fi




%% file: main_arXiv.bbl
\begin{thebibliography}{2}
    \bibitem[Huang et~al.(2025)]{huang2025r}
    Chengsong Huang et~al.
    \newblock R-Zero: Self-Evolving Reasoning LLM from Zero Data.
    \newblock \emph{arXiv preprint arXiv:2508.05004}, 2025.
    \newblock \url{https://arxiv.org/abs/2508.05004}.
    \bibitem[Xia et~al.(2025)]{xia2025agent0}
    Peng Xia et~al.
    \newblock Agent0: Unleashing Self-Evolving Agents from Zero Data via
    Tool-Integrated Reasoning.
    \newblock \emph{arXiv preprint arXiv:2511.16043}, 2025.
    \newblock \url{https://arxiv.org/abs/2511.16043}.
    \end{thebibliography}
